\documentclass{IOS-Book-Article}

\usepackage{mathptmx}
\usepackage{amsmath}
\usepackage{amssymb}
\usepackage{algorithm,algpseudocode}
\usepackage{multirow}
\usepackage{xcolor}

\usepackage{graphicx}
\usepackage{amsthm}
\newtheorem{theorem}{Theorem}[section]

\theoremstyle{definition}
\newtheorem{definition}[theorem]{Definition}
\theoremstyle{remark}

\usepackage{booktabs}
\usepackage{caption}
\usepackage{threeparttable}
\usepackage{tabularx}
\usepackage{colortbl}
\definecolor{mygray}{gray}{.9}
\usepackage{cite}
\usepackage{epstopdf}
\usepackage{url}
\usepackage[pdftex,linkcolor=blue,citecolor=blue,backref=page]{hyperref}
\usepackage{float}
\usepackage[toc,page]{appendix}

\usepackage[utf8]{inputenc}
\usepackage[english]{babel}
\usepackage{fancyhdr}
\usepackage{afterpage}

\def\hb{\hbox to 10.7 cm{}}

\begin{document}
\thispagestyle{headings}
\setcounter{page}{1}
\pagenumbering{arabic}
\newpage
\pagestyle{plain}
\setcounter{page}{1}
\pagenumbering{arabic}
\pagestyle{fancy}
\fancyhf{}
\chead{T. Mo et al. / An influence-based fast preceding questionnaire model}
\fancyhead[R]{\thepage}
\renewcommand{\headrulewidth}{0pt}
\def\thepage{}

\begin{frontmatter} 

\title{An influence-based fast preceding questionnaire model for elderly assessments}

\markboth{}{November 2016\hb}

\author[add1]{\fnms{Tong Mo}
\thanks{Corresponding Author: Tong Mo, School of Software and Microelectronics, Peking University, Beijing 100086, China; E-mail: motong@ss.pku.edu.cn}},
\author[add1]{\fnms{Rong Zhang}
\thanks{Rong Zhang, School of Software and Microelectronics, Peking University, Beijing 100086, China; E-mail: rongzhanghappy@pku.edu.cn}},
\author[add1]{\fnms{Weiping Li}
\thanks{Weiping Li, School of Software and Microelectronics, Peking University, Beijing 100086, China; E-mail: wpli@ss.pku.edu.cn}},
\author[add1]{\fnms{Jingbo Zhang}
\thanks{Jingbo Zhang, School of Software and Microelectronics, Peking University, Beijing 100086, China; E-mail: 35244@qq.com}},
\author[add1]{\fnms{Zhonghai Wu}
\thanks{Zhonghai Wu, School of Software and Microelectronics, Peking University, Beijing 100086, China; E-mail: wuzh@pku.edu.cn}}
and
\author[add2]{\fnms{Wei Tan}
\thanks{Wei Tan, IBM T. J. Watson Research Center, Yorktown Heights, NY, USA; E-mail: wtan@us.ibm.com}}

\runningauthor{B.P. Manager et al.}
\address[add1]{School of Software and Microelectronics, Peking University, Beijing, China}
\address[add2]{IBM T. J. Watson Research Center, Yorktown Heights, NY, USA}

\begin{abstract}
To improve the efficiency of elderly assessments, an influence-based fast preceding questionnaire model (FPQM) is proposed. Compared with traditional assessments, the FPQM optimizes questionnaires by reordering their attributes. The values of low-ranking attributes can be predicted by the values of the high-ranking attributes. Therefore, the number of attributes can be reduced without redesigning the questionnaires. A new function for calculating the influence of the attributes is proposed based on probability theory. Reordering and reducing algorithms are given based on the attributes' influences. The model is verified through a practical application. The practice in an elderly-care company shows that the FPQM can reduce the number of attributes by 90.56\% with a prediction accuracy of 98.39\%. Compared with other methods, such as the Expert Knowledge, Rough Set and C4.5 methods, the FPQM achieves the best performance. In addition, the FPQM can also be applied to other questionnaires.
\end{abstract}

\end{frontmatter}
\markboth{November 2016\hb}{November 2016\hb}

\section{Introduction}

Questionnaires have been widely used in various fields, including elderly assessments. Several questionnaires have been developed and are currently in extensive use to assess health-related quality of life (HRQOL)\cite{mcdowell2006measuring}. Aging is an increasingly serious social phenomenon in China, and there is a strong need for care services. Assessments of the elderly are essential for providing personalized services. Existing assessment methods are usually based on the Barthel Index\cite{mahoney1965functional} and the national industry standard for the ability assessment of elderly adults\cite{wang2014ability}. Many investigation attributes are needed to systematically obtain  information about the elderly. The elderly are asked about multiple attributes in succession. These assessment methods are inefficient, and the order of the attributes is not reasonable. When there is a relationship between attributes, some unknown attributes can be predicted by known attributes, and a more reasonable order should be determined\cite{drummond2008questionnaire,faulkner1990measures,mccoll2003generic}.

Classical Test Theory (CTT), Rasch Analysis (RA), decision rule, and experts\cite{prieto2003classical,fernandez2001affective,kitisomprayoonkul2006thai,rosen1999development,badia2010bone,nijsten2006testing} have been applied to reduce the length of health questionnaires. However, actually, these removed attributes have provided additional information. A reasonable order of these attributes of the questionnaires can also be considered. Correlation, multiple regression, factor analysis, cluster analysis and structural equation modelling, and hierarchical multiple regression \cite{dima2010living,arnow2006comorbid,rippentrop2005relationship,vines2003relationship} can be used to determine the relationships among the attributes of health questionnaires. Certain attributes can indeed be predicted by other attributes using hierarchical logistic regression, correlation analysis, and binary logistic stepwise regression\cite{kersh2001psychosocial,aoyama2011physical,aydeniz2015physical,gatz2005depressive,cruice2005personal}. However, only one attribute, not multiple attributes simultaneously, could be predicted in one study. Meanwhile, the involved attributes in each study are incomplete.

A solid mathematical definition of the question is given. The fast preceding questionnaire model (FPQM) is proposed to solve the problem in five steps. First, the influence of one attribute on all other attributes is defined and calculated. Second, we traverse every investigation attribute and chooses the attribute  with  the largest influence as the best attribute to split. Third, we create the FPQM with the best attribute. We traverse every value of the attribute, and the sub-dataset corresponding to the value can be used to obtain the sub-model recursively. Then, the sub-model is attached to the full model, and the full model is obtained when the recursion ends. Fourth, the created FPQM is used for the real investigation. The value is directly asked for at the beginning of the real investigation because there is no prior information about the respondent. Certain investigation attributes can be inferred after sufficient information has been accumulated. At that time, the confidence level is greater than the given threshold, which means that questions about the attribute do not need to be asked. Fifth, we calculate the evaluation metrics and evaluate the model FPQM.

This paper is organized as follows. Section \ref{related work} reviews related work. The fast preceding questionnaire model (FPQM) is introduced in Section \ref{fpqm}. First, a solid mathematical definition of the question is given. Then, an influence calculation formula, the best attribute to split choosing algorithm (BASCA), the fast preceding questionnaire model creating algorithm (FPQMCA), the model used for real investigation algorithm (MURIA), and the model evaluation algorithm (MEA) are presented. Section \ref{experiments} shows the experimental results, therein presenting the experimental data;, evaluation metrics; the overall results of the FPQM; the comparison experiment with Expert Knowledge, Rough Set, and C4.5; and the factor analysis, which includes the number of elderly, number of investigation attributes, and threshold. Section \ref{conclusion} concludes the paper.

\section{Related work}
\label{related work}
\subsection{Attribute reduction}

Luis Prieto \cite{prieto2003classical} presents a parallel reduction in a 38-attribute questionnaire, the Nottingham Health Profile (NHP), to empirically compare Classical Test Theory (CTT) and Rasch Analysis (RA) results. The CTT results in 20 attributes (4 dimensions), whereas RA results in 22 attributes (2 dimensions). Moreover, the attribute-total correlation ranges from 0.45-0.75 for NHP20 and from 0.46-0.68 for NHP22, while the reliability ranges from 0.82-0.93 and from 0.87-94, respectively.

Ephrem Fernandez \cite{fernandez2001affective} reduces and reorganizes the McGill Pain Questionnaire (MPQ) using a 3-step decision rule for affective and evaluative descriptors of Pain. With a minimum absolute frequency of 17 and a minimum relative frequency of 1/2 as the threshold values, the words of the MPQ are reduced from 78 to less than 20 on average. This reduction leads to a negligible loss of information transmitted. Moreover, Wasuwat Kitisomprayoonkul \cite{kitisomprayoonkul2006thai} develops the Thai Short-Form McGill Pain Questionnaire (Th-SFMPQ).

RC Rosen \cite{rosen1999development} develops an abridged five-attribute version (IIEF-5) of the 15-attribute International Index of Erectile Function (IIEF) to diagnose the presence and severity of erectile dysfunction (ED). The five attributes are selected based on the ability to identify the presence or absence of ED and on adherence to the National Institute of Health's definition of ED. The IIEF-5 possesses favorable properties for detecting the presence and severity of ED.

X. Badia \cite{badia2010bone} achieves a qualitative and quantitative reduction in the 179 expressions of the bone metastasis quality of life questionnaire (BOMET-QOL) with respect to clarity, frequency and importance with 15 experts. This phase, which is performed in two steps, results in the 35-attribute version of the BOMET-QOL. The initial reduction yields a 25-attribute questionnaire via factorial analysis. Similarly, the BOMET-QOL-25 is reduced to an integrated version of 10 attributes through a sample of 263 oncology patients. The BOMET-QOL is an accurate, reliable and precise 10-attribute instrument for assessing HRQOL.

Tamar E.C. Nijsten \cite{nijsten2006testing} tests and reduces Skindex-29 to Skindex-17 using Rasch Analysis. The Rasch Analysis of the combined emotion and social functioning subscale of Skindex-29 results in a 12-attribute psychosocial subscale. A total of five of the seven attributes are retained in a symptom subscale. Classical psychometric properties, such as the response distribution, attribute–rest correlation, attribute complexity, and internal consistency, of the two subscales of Skindex-17 are at least adequate. Skindex-17 is a Rasch-reduced version of Skindex-29, with two independent scores that can be used for the measurement of health-related quality of life (HRQOL) for dermatological patients.

\cite{prieto2003classical,fernandez2001affective,kitisomprayoonkul2006thai,rosen1999development,badia2010bone,nijsten2006testing} remove some attributes directly and develop qualitative and quantitative reductions in questionnaires about health using Classical Test Theory (CTT), Rasch Analysis (RA), decision rules, or experts. These questionnaires include the Nottingham Health Profile (NHP), McGill Pain Questionnaire (MPQ), 15-attribute International Index of Erectile Function (IIEF), bone metastasis quality of life questionnaire (BOMET-QOL), and Skindex-29. However, these removed attributes can provide additional information, and their values can be predicted by the remaining attributes with reduction methods. Meanwhile, a more reasonable order of these attributes is not considered.

\subsection{Relationships among attributes}

Alexandra-Lelia Dima \cite{dima2010living} studies the interrelations between acceptance, emotions, illness perceptions and health status. The confirmatory analysis (employing a variety of statistical procedures, from correlation to multiple regression, factor analysis, cluster analysis and structural equation modelling) largely confirms the expected relations within and between domains and is also informative regarding the most suitable data reduction methods. An additional exploratory analysis focuses on identifying the comparative characteristics of acceptance, emotions, and illness perceptions in predicting health status metrics.

Arnow, Bruce A \cite{arnow2006comorbid} provides estimates of the prevalence and strength of association between major depression and chronic pain in a primary care population and examines the clinical burden associated with the two conditions alone and together. Data are collected by questionnaires assessing major depressive disorder (MDD), chronic pain, pain-related disability, somatic symptom severity, panic disorder, other anxiety, probable alcohol abuse, and health-related quality of life (HRQL). The instruments include the Patient Health Questionnaire, SF-8, and the Graded Chronic Pain Questionnaire. The conclusions are that chronic pain is common among those with MDD, and Comorbid MDD and disabling chronic pain are associated with greater clinical burden than is MDD alone.

A. Elizabeth Rippentrop \cite{rippentrop2005relationship} seeks to better understand the relationships among religion/spirituality and physical health, mental health, and pain in 122 patients with chronic musculoskeletal pain. Hierarchical multiple regression analyzes reveal significant associations between components of religion/spirituality and physical and mental health. Forgiveness, negative religious coping, daily spiritual experiences, religious support, and self-rankings of religious/spiritual intensity significantly predict mental health status. Religion/spirituality is unrelated to pain intensity and life interference due to pain. Religion/spirituality may have both costs and benefits for the health of those with chronic pain.

Susan W. Vines \cite{vines2003relationship} determines the relationships between pain perceptions, immune function, depression and health behaviors and examines the effects of chronic pain on immune function using depression and health behaviors as covariates. Pain perceptions show positive significant correlations with depression (P = 0.01) and total percent of NK cells (P = 0.04). Depression and health behaviors are negatively correlated (P = 0.01). Positive associations are observed for depression and 2 PHA mitogen levels (P$<$0.05).  The immune function of patients with chronic pain is significantly higher than in the no-pain comparison group. Pain perceptions may have a deleterious effect on enumerative NK cell measures and depression levels.

Most of the attributes mentioned in \cite{dima2010living,arnow2006comorbid,rippentrop2005relationship,vines2003relationship} are included in Table \ref{tab:Assessattributes}, such as acceptance, emotions, illness perceptions, and health status; depression, chronic pain, and clinical burden; religion/spirituality and physical health, mental health, and pain; and pain perceptions, immune function, depression and health behaviors. The only two differences between the investigation attributes in Table \ref{tab:Assessattributes} and the attributes mentioned in the literature are the expressions. The attributes mentioned in the literature are more conceptual. Applied methods include correlation, multiple regression, factor analysis, cluster analysis, structural equation modelling, and hierarchical multiple regression. The literature proves that relationships among these attributes do exist. However, the attributes covered by the relationships in each study are incomplete, and the relationships have not been well utilized to provide results of interest such as in prediction.

\subsection{Prediction}

Kersh, BC \cite{kersh2001psychosocial} uses psychosocial and health status variables independently to predict health care seeking for fibromyalgia. Subjects are administered 14 measures, which produce six domains of variables: background demographics and pain duration; psychiatric morbidity; and personality, environmental, cognitive, and health status factors. These domains are input into 4 different hierarchical logistic regression analyzes to predict the status as patient or non-patient. The full regression model is statistically significant (P$<$0.0001) and correctly identifies 90.7\% of the subjects, with a sensitivity of 92.4\% and a specificity of 87.2\%.

Maki Aoyama \cite{aoyama2011physical} uses physical and functional factors in activities of daily living to predict falls in community-dwelling older women. Correlation analysis investigating associations among the scores of assessment scales and actual measurements of muscle strength and balance shows that there are significant correlations between handgrip strength and the Falls Efficacy Scale, Functional Reach test, Timed Up and Go test, Berg Balance Scale, Motor Fitness Scale, and Motor Functional Independence Measure in fallers and non-fallers. A binary logistic stepwise regression analysis reveals that only an inability of “being able to go up and down the staircase” in the Motor Fitness Scale remains a significant variable to predict falls.

Aydeniz, Ali \cite{aydeniz2015physical} predicts falls in the elderly with physical, functional and sociocultural parameters. Falls are common in patients with weakness, fatigue, dizziness, and swelling in the legs and in subjects with appetite loss. Fallers have lower functional status than do non-fallers (p=0.028). In addition, fallers have more depressive symptoms than do non-fallers (p=0.019). Quality of life (NHP), especially physical activity, energy level and emotional reaction, subgroups are different (p=0.016, 0.015, and 0.005, respectively). Disability and mental status are similar in groups (p=0.006). Musculoskeletal problems, functional status and social status might be contributors to falls.

Jennifer L. Gatz \cite{gatz2005depressive} uses depressive symptoms to predict Alzheimer's disease and dementia. The Total Center for Epidemiologic Studies Depression (CES-D) score is a significant predictor of AD and dementia when categorized as a dichotomous variable according to the cutoff scores of 16 and 17; a CES-D cutoff of 21 is a significant predictor of AD and a marginally significant predictor of dementia. When analyzed as a continuous variable, the CES-D score is marginally predictive of AD and dementia. Neither participant-reported history of depression nor participant-reported duration of depression is significant in predicting AD or dementia. 

Madeline Cruice \cite{cruice2005personal} predicts social participation in older adults with personal factors, communication and vision. Assessments are individually conducted in a face-to-face interview situation with the primary researcher, who is a speech pathologist. Social participation is shown to be associated with vision, communication activities, age, education and emotional health. Naming and hearing impairments are not reliable predictors of social participation. It is concluded that professionals interested in maintaining and improving the social participation of older people should strongly consider these predictors in community-directed interventions.

Most of the attributes mentioned in \cite{kersh2001psychosocial,aoyama2011physical,aydeniz2015physical,gatz2005depressive,cruice2005personal} are also included in Table \ref{tab:Assessattributes} such as psychosocial and health status variables and health care; physical, functional and sociocultural parameters and falls; depressive symptoms, Alzheimer's disease and dementia; and personal factors, communication and vision, and social participation. Health care, falls, Alzheimer's disease, dementia, and social participation are predicted using hierarchical logistic regression, correlation analysis, and binary logistic stepwise regression, respectively. The literature here is illustrative of the fact that certain attributes can indeed be predicted by other attributes. However, one study only predicts one attribute, not multiple attributes simultaneously. Meanwhile, the involved attributes in each study are also incomplete. A sufficient prediction between complete attributes can be studied.

The relation and motivation between all the related works and this research are emphasized and explained here. \cite{prieto2003classical,fernandez2001affective,kitisomprayoonkul2006thai,rosen1999development,badia2010bone,nijsten2006testing} show that certain attributes in Table \ref{tab:Assessattributes} can be reduced directly. The information of these attributes is redundant and contained in other attributes. \cite{dima2010living,arnow2006comorbid,rippentrop2005relationship,vines2003relationship} provide further evidence that there is an inherent relationship between these attributes. \cite{kersh2001psychosocial,aoyama2011physical,aydeniz2015physical,gatz2005depressive,cruice2005personal} further show that certain attributes can be predicted by other attributes because of the underlying relationship. All the related works form the foundation of this research. The proposed fast preceding questionnaire model (FPQM) can achieve state-of-the-art performance only when there is an inherent relationship between attributes. If the relationship does not exist at all, no methods can predict certain attributes by others. The relationship is the foundation of all possible methods, including the FPQM. \cite{prieto2003classical,fernandez2001affective,kitisomprayoonkul2006thai,rosen1999development,badia2010bone,nijsten2006testing} reduce some attributes directly, while the FPQM predicts the values of the attributes and preserves these attributes. In addition, In\cite{kersh2001psychosocial,aoyama2011physical,aydeniz2015physical,gatz2005depressive,cruice2005personal}, one study only predicts one attribute, while the FPQM can predict multiple attributes simultaneously.

\section{Fast preceding questionnaire model (FPQM)}
\label{fpqm}

A solid mathematical definition of question is given, and the fast preceding questionnaire model (FPQM) is proposed to solve the problem in five steps.

\textbf{Step 1: Calculate the influence}

We calculate, in order, the confidence level of the attribute by taking a value under the condition of another attribute taking a value, the influence of the attribute taking a value on another attribute, the influence of the attribute on another attribute, the influence of the attribute on all other attributes, and the attribute that has the largest influence on all other attributes.

\textbf{Step 2: Choose the best attribute to split}

We traverse every investigation attribute, every other attribute, every value of the attribute, and every value of the other attribute to calculate every influence. Finally, the influence of the investigation attribute on all other attributes can be calculated. Then, we  logically choose the best attribute that has the largest influence on all other attributes.

\textbf{Step 3: Create the FPQM}

After the best attribute to split is chosen, we traverse every value of the attribute, and the sub-model can be obtained with sub-dataset corresponding to the value recursively. Then, we attach the sub-model to the full model, and the full  model is obtained when the recursion ends.

\textbf{Step 4: Use the FPQM for real investigation}

Now, the FPQM can be used to investigate a new respondent. At the beginning of the real investigation, there is no prior information about the respondent; therefore, we ask for the value directly. After sufficient information has been accumulated, some investigation attributes can be inferred. If the confidence level is larger than the given threshold, then the attribute does not need to be asked about.

\textbf{Step 5: Evaluate the model}

After the FPQM is used to investigate the new respondent, the evaluation metrics can be calculated. The FPQM can be evaluated based on these metrics.

\begin{figure}[ht]
    \centering
    \includegraphics[width=0.25\textwidth]{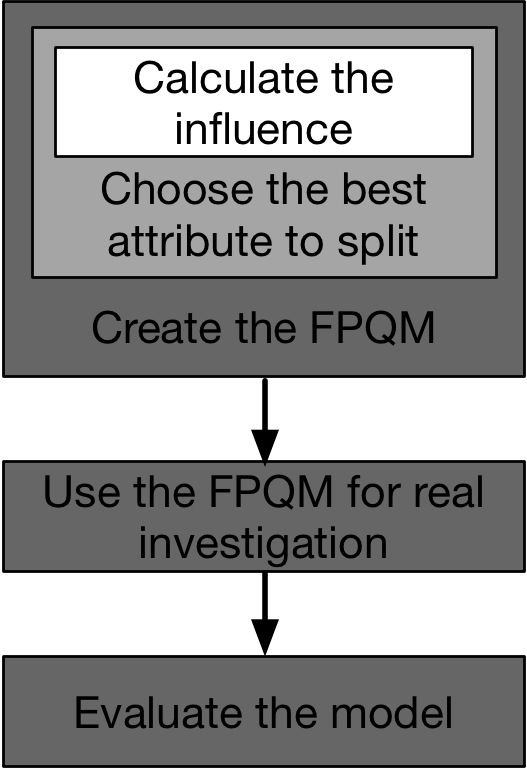}
    \caption{Illustration of the five steps.}
    \label{fig:graph}
\end{figure}

The five steps are also presented in the form of a graph, as shown in Figure \ref{fig:graph}. Step 1: Calculate the influence with Eq.~\eqref{eq1}-\eqref{eq5}. Step 2: Choose the best attribute to split with Algorithm  \ref{BASCA} and the  calculated influence in Step 1. Step 3: Create the FPQM with Algorithm  \ref{FPQMCA}, and at every recursion step, call Algorithm  \ref{BASCA} to choose the best attribute to split. Step 4: Use the FPQM for the real investigation with Algorithm  \ref{MURIA} after the FPQM is created in Step 3. Step 5: Evaluate the model with Algorithm  \ref{MEA}.

\subsection{Problem definition}

  \begin{definition}
  \label{def1}
    Let $U=\{U_1,U_2,...,U_m\}$ be the collection of all individuals who are investigated in the training dataset, where $U_i (1\leqslant i \leqslant m)$ is the $i$-th individual being investigated.
  \end{definition}

\begin{definition}
    Let $\tilde{U}=\{\tilde{U_1},\tilde{U_2},...,\tilde{U_{\tilde{m}}}\}$ be the collection of all individuals in the testing dataset, where $\tilde{U_i} (1\leqslant i \leqslant \tilde{m})$ is the $i$-th individual and $\tilde{U}\cap U=\phi $.
  \end{definition}

\begin{definition}
\label{V}
    Let $V=\{V_1,V_2,...,V_n\}$ be the collection of investigation attributes, where $V_j$ is the $j$-th investigation attribute.
  \end{definition}

\begin{definition}
\label{W}
    Let $W=\{W_1,W_2,...,W_n\}$ be the collection of all possible values on all investigation attributes, where $W_j$ is the collection of all possible values on $V_j$.
  \end{definition}

\begin{definition}
\label{N}
    $N=\{N_1,N_2,...,N_n\}$, where $N_j$ is the number of $W_j$. $N$ will be used in the time complexity analysis of the following four algorithms.
  \end{definition}
  
\begin{definition}
    Let $D=\{D_{ij}\} (1\leqslant i\leqslant m, 1\leqslant j\leqslant n)$ be the matrix of all real values of all individuals, and $D$ is the training dataset, where $D_{ij}$ is the real value of individual $U_i$ on the investigation attribute $V_j$ and $D_{ij}\in W_j$.
  \end{definition}

\begin{definition}
    Let $R=\{R_{ij}\} (1\leqslant i\leqslant \tilde{m}, 1\leqslant j\leqslant n)$ be the matrix of all real values, where $R$ is the testing dataset, in which $R_{ij}$ is the real value of $\tilde{U_i}$ on $V_j$ and $R_{ij}\in W_j$. $R$ can also be represented as $R=\{R_i\} (1\leqslant i\leqslant \tilde{m})$; $R_i=\{R_{ij}\} (1\leqslant j\leqslant n)$, where $R_i$ is the vector of real values of $\tilde{U_i}$.
  \end{definition}

\begin{definition}
\label{R'}
    Let $R'=\{R_{ij}'\} (1\leqslant i\leqslant \tilde{m}, 1\leqslant j\leqslant n)$ be the matrix of all final values, where $R_{ij}'$ is the final value of $\tilde{U_i}$ on $V_j$ and $R_{ij}'\in W_j$. $R'$ can also be represented as $R'=\{R_{ij}'\}(1\leqslant i\leqslant \tilde{m})$; $R_i'=\{R_{ij}'\} (1\leqslant j\leqslant n)$, where $R_i'$ is the vector of the final values of $\tilde{U_i}$.
  \end{definition}

\begin{definition}
    Let $K=\{K_{ij}\} (1\leqslant i\leqslant \tilde{m}, 1\leqslant j\leqslant n)$ be the matrix of indication values showing whether the final value $R_{ij}'$ is equal to the real value $R_{ij}$. $K_{ij}\in \{0,1\}$, where $K_{ij}=1$ when $R_{ij}'=R_{ij}$ and $K_{ij}=0$ when $R_{ij}'\neq R_{ij}$.
  \end{definition}

\begin{definition}
\label{P}
    Let $P=\{P_{ij}\} (1\leqslant i\leqslant \tilde{m}, 1\leqslant j\leqslant n)$ be the matrix of the confidence levels of $\tilde{U_i}$ taking the final values $R_{ij}'$. $P_{ij}\in [0,1]$, where $P_{ij}$ is the confidence level of $\tilde{U_i}$ taking $R_{ij}'$ on $V_j$. $P$ can also be represented as $P=\{P_i\} (1\leqslant i\leqslant \tilde{m})$; $P_i=\{P_{ij}\} (1\leqslant j\leqslant n)$, where $P_i$ is the vector of the confidence levels of $\tilde{U_i}$.
  \end{definition}

\begin{definition}
\label{I}
    Let $I=\{I_{ij}\} (1\leqslant i\leqslant \tilde{m}, 1\leqslant j\leqslant n)$ be the matrix of indication values showing whether the value $R_{ij}'$ is predicted from other already known attributes. $I_{ij}\in \{0,1\}$, where $I_{ij}=1$ denotes that $R_{ij}'$ is predicted and $I_{ij}=0$ denotes that $R_{ij}'=R_{ij}$ by asking $\tilde{U_i}$ on $V_j$ directly. Note that $I_{ij}=1$ when $P_{ij}\geqslant \sigma $, and $I_{ij}=0$ when $P_{ij}< \sigma $, where $\sigma $ is the given threshold. $I$ can also be represented as $I=\{I_i\}(1\leqslant i\leqslant \tilde{m})$; $I_i=\{I_{ij}\} (1\leqslant j\leqslant n)$, where $I_i$ is the vector of the indication values of $\tilde{U_i}$.
  \end{definition}

\begin{definition}
\label{O}
    Let $O=\{O_1,O_2,...,O_{\tilde{m}}\}$ be the collection of all reasonable orders in which individuals should be investigated, where $O_i$ is a reasonable order in which $\tilde{U_i}$ should be investigated. $O_i=\{j_1,j_2,...,j_n\}$ is a single substitution of $\{1,2,...,n\}$. $O_{i_1}$ is different from $O_{i_2}$ (which it most likely is) when $i_1$ is different from $i_2$.
  \end{definition}

\begin{definition}
    Let $M$ be the fast preceding questionnaire model. $M$ is a tree structure, and $M$ will determine $R'$, $I$, and $O$.
  \end{definition}

\begin{definition}
    Let $C$ be the collection of all appearing confidence levels when creating $M$, where $C$ will determine $P$.
  \end{definition}

\begin{definition}
    Let $S=\{U,V,W,D,R,R',K,P,I,O,M,C\}$ be the system of the fast preceding questionnaire model.
  \end{definition}

\begin{definition}
    Let $\bar{S}$ be the space of all possible questionnaire models.
  \end{definition}

\begin{definition}
    Average accuracy rate ($AAR$): $AAR=\tfrac{1}{\tilde{m}}\sum_{i=1}^{\tilde{m}}\tfrac{\sum_{j=1}^{n}K_{ij}*I_{ij}}{\sum_{j=1}^nI_{ij}}$. $AAR$ describes how accurate the model can be. $AAR$ can also be represented as $AAR=\tfrac{1}{\tilde{m}}\sum_{i=1}^{\tilde{m}}AR_i$; $AR_i=\tfrac{\sum_{j=1}^{n}K_{ij}*I_{ij}}{\sum_{j=1}^nI_{ij}}$, where $AR_i$ is the accuracy rate of $\tilde{U_i}$.
  \end{definition}

\begin{definition}
    Average reduction rate ($ARR$): $ARR=\tfrac{1}{\tilde{m}}\sum_{i=1}^{\tilde{m}}\tfrac{\sum_{j=1}^nI_{ij}}{n}$. $ARR$ describes how well the model can accelerate the questionnaire. $ARR$ can also be represented as $ARR=\tfrac{1}{\tilde{m}}\sum_{i=1}^{\tilde{m}}RR_i$; $RR_i=\tfrac{\sum_{j=1}^nI_{ij}}{n}$, where $RR_i$ is the reduction rate $\tilde{U_i}$.
  \end{definition}

\begin{definition}
    Average $F_\beta $-Measure: $AF_\beta =\tfrac{1}{\tilde{m}}\sum_{i=1}^{\tilde{m}}\tfrac{(\beta ^2+1)*AR_i*RR_i}{\beta ^2*AR_i+RR_i}$, where $\beta $ is a given parameter. $AF_\beta $ describes a balance between $AAR$ and $ARR$.
  \end{definition}

\begin{definition}
    The problem is defined as the following.
    \begin{flalign}
        max\ AF_\beta \ \ subject\ to\ S\in \bar{S}  \nonumber
    \end{flalign}
  \end{definition}

\subsection{Influence calculation formula}

To create the model from the training dataset $D$, the influence of one investigation attribute on all others should be calculated. The influence calculation formula is given in Definition \ref{d1}-\ref{d5} when the depth of the created model reaches t. The influence of the investigation attribute depends on the influence of the values.

\begin{definition}
\label{d1}
    The confidence level of the investigation attribute $V_{j_{k_2}}^t$ taking the value $v_{j_{k_2}}^t$ under the condition of the investigation attribute $V_{j_{k_1}}^t$ taking the value $v_{j_{k_1}}^t$ when the previous k-1 layer values $V^1,V^2,...,V^{t-1}$ are already known.
  \end{definition}

\begin{align}
\label{eq1}
&P(V_{j_{k_2}}^t=v_{j_{k_2}}^t|V_{j_{k_1}}^t=v_{j_{k_1}};V^1,V^2,...,V^{t-1})\nonumber\\
&=\tfrac{N(V_{j_{k_2}}^t=v_{j_{k_2}}^t,V_{j_{k_1}}^t=v_{j_{k_1}};V^1,V^2,...,V^{t-1})}{N(V_{j_{k_1}}^t=v_{j_{k_1}}^t;V^1,V^2,...,V^{t-1})}
\end{align}

where $V_{j_{k_1}}^t\in V$, $V_{j_{k_2}}^t\in V$, $v_{j_{k_1}}^t\in W_{j_{k_1}}$, and $v_{j_{k_2}}^t\in W_{j_{k_2}}$. $V^1$ is the $1st$-layer value of all investigation attributes $V$, $V^2$ is the $2nd$ layer value of $V$, etc. $N(V_{j_{k_2}}^t=v_{j_{k_2}}^t,V_{j_{k_1}}^t=v_{j_{k_1}};V^1,V^2,...,V^{t-1})$ is the number of individuals in $U$ who take the value $v_{j_{k_1}}^t$, $v_{j_{k_2}}^t$ on the investigation attributes $V_{j_{k_1}}^t$, $V_{j_{k_2}}^t$, respectively. $N(V_{j_{k_1}}^t=v_{j_{k_1}}^t;V^1,V^2,...,V^{t-1})$ is the number of individuals in $U$ who take the value $v_{j_{k_1}}^t$ on $V_{j_{k_1}}^t$.

\begin{definition}
\label{d2}
    The influence of $V_{j_{k_1}}^t$ taking $v_{j_{k_1}}^t$ on $V_{j_{k_2}}^t$ when the previous k-1 layer values $V^1,V^2,...,V^{t-1}$ are already known.
  \end{definition}

\begin{align}
\label{eq2}
&INF(V_{j_{k_2}}^t|V_{j_{k_1}}^t=v_{j_{k_1}};V^1,V^2,...,V^{t-1})\nonumber\\
&=\sum_{v_{j_{k{}_2}}^t\in W_{j_{k_2}}}P(V_{j_{k_2}}^t=v_{j_{k_2}}^t|V_{j_{k_1}}^t=v_{j_{k_1}};V^1,V^2,...,V^{t-1})^2
\end{align}

where $P(V_{j_{k_2}}^t=v_{j_{k_2}}^t|V_{j_{k_1}}^t=v_{j_{k_1}};V^1,V^2,...,V^{t-1})$ is defined in Definition \ref{d1}. Notice that $INF(V_{j_{k_2}}^t|V_{j_{k_1}}^t=v_{j_{k_1}};V^1,V^2,...,V^{t-1})$ is not defined as  $INF(V_{j_{k_2}}^t|V_{j_{k_1}}^t=v_{j_{k_1}};V^1,V^2,...,V^{t-1})=\sum_{v_{j_{k{}_2}}^t\in W_{j_{k_2}}}P(V_{j_{k_2}}^t=v_{j_{k_2}}^t|V_{j_{k_1}}^t=v_{j_{k_1}};V^1,V^2,...,V^{t-1})$
 because $\sum_{v_{j_{k{}_2}}^t\in W_{j_{k_2}}}P(V_{j_{k_2}}^t=v_{j_{k_2}}^t|V_{j_{k_1}}^t=v_{j_{k_1}};V^1,V^2,...,V^{t-1})=1$ is always true.

\begin{definition}
\label{d3}
    The influences of $V_{j_{k_1}}^t$ on $V_{j_{k_2}}^t$ when the previous k-1 layer values $V^1,V^2,...,V^{t-1}$ are already known.
  \end{definition}

\begin{align}
\label{eq3}
&INF(V_{j_{k_2}}^t|V_{j_{k_1}}^t;V^1,V^2,...,V^{t-1})\nonumber\\
&=\sum_{v_{j_{k_1}}^t\in W_{j_{k_1}}}\biggl[P(V_{j_{k_1}}^t=v_{j_{k_1}}^t;V^1,V^2,...,V^{t-1})\ast\nonumber\\
&INF(V_{j_{k_2}}^t|V_{j_{k_1}}^t=v_{j_{k_1}}^t;V^1,V^2,...,V^{t-1})\biggr]
\end{align}

where $P(V_{j_{k_1}}^t=v_{j_{k_1}}^t;V^1,V^2,...,V^{t-1})$ is the confidence level of $V_{j_{k_1}}^t$ taking $v_{j_{k_1}}^t$.

\begin{definition}
\label{d4}
    The influence of $V_{j_{k_2}}^t$ on all other investigation attributes $\Delta $ when the previous k-1 layer values $V^1,V^2,...,V^{t-1}$ are already known.
  \end{definition}

\begin{align}
\label{eq4}
&INF(\Delta |V_{j_{k_1}}^t;V^1,V^2,...,V^{t-1})\nonumber\\
&=\sum_{V_{j_{k_2}}^t\in V, k_2\neq k_1}INF(V_{j_{k_2}}^t|V_{j_{k_1}}^t;V^1,V^2,...,V^{t-1})^2
\end{align}

where $\Delta =V-\{V_{j_{k_1}}^t\}$.

\begin{definition}
\label{d5}
    The investigation attribute $V_*^t$ that has the largest influence on all other investigation attributes $\Delta $.
  \end{definition}

\begin{equation}
\label{eq5}
V_*^t=\underset{V_{j_{k_1}}^t\in V}{arg\ max }\ INF(\Delta |V_{j_{k_1}}^t;V^1,V^2,...,V^{t-1})
\end{equation}

\subsection{Best attribute to split choosing algorithm (BASCA)}

When creating the FPQM, it is necessary to choose the best attribute to split. When the depth of the created model reaches $t$, we traverse every investigation attribute; then, we traverse every other investigation attribute and calculate the influence of the investigation attribute on all other attributes. Lines \ref{BASCA_L5}, \ref{BASCA_L7}, \ref{BASCA_L8}, \ref{BASCA_L9}, and \ref{BASCA_L10} are calculated with Eqs~\eqref{eq1}-\eqref{eq5}. Then, we logically choose $V_*^t$ as the best attribute that has the largest influence on $\Delta $. The pseudocode is shown in Algorithm \ref{BASCA} when the depth of the created model reaches $t$.

\begin{algorithm}[h!]
\caption{Best attribute to split choosing algorithm (BASCA)
\newline \textbf{Input:} $D$, $V$
\newline \textbf{Output:} $V_*^t$
}
\label{BASCA}
\begin{algorithmic}[1]
\For{($V_{j_{k_1}}^t\in V$)}
\label{BASCA_L1}
\For{($V_{j_{k_2}}^t\in V$ \& $V_{j_{k_1}}^t\neq V_{j_{k_2}}^t$)}
\label{BASCA_L2}
\For{each value $v_{j_{k_1}}^t$ of $V_{j_{k_1}}^t$}
\label{BASCA_L3}
\For{each value $v_{j_{k_2}}^t$ of $V_{j_{k_2}}^t$}
\label{BASCA_L4}
\State Calculate $P(v_{j_{k_2}}^t|v_{j_{k_1}}^t;V^1,V^2,...,V^{t-1})$
\label{BASCA_L5}
\State Add $P(v_{j_{k_2}}^t|v_{j_{k_1}}^t;V^1,V^2,...,V^{t-1})$ to $C$
\label{BASCA_L6}
\EndFor
\State Calculate $INF(V_{j_{k_2}}^t|v_{j_{k_1}}^t;V^1,V^2,...,V^{t-1})$
\EndFor
\label{BASCA_L7}
\State Calculate $INF(V_{j_{k_2}}^t|V_{j_{k_1}}^t;V^1,V^2,...,V^{t-1})$
\EndFor
\label{BASCA_L8}
\State Calculate $INF(\Delta |V_{j_{k_1}}^t;V^1,V^2,...,V^{t-1})$
\EndFor
\label{BASCA_L9}
\State Calculate $V_*^t$
\label{BASCA_L10}
\State \Return $V_*^t$
\end{algorithmic}
\end{algorithm}

$V_*^t= BASCA(D,V)$. The BASCA returns the investigation attribute $V_*^t$ that has the largest influence on $\Delta $. $C$ is a global variable, and the BASCA can also be called to obtain $C$.

\subsection{Fast preceding questionnaire model creating algorithm (FPQMCA)}

Now, the FPQM can be created with the above groundwork. After $V_*$ is chosen as the best attribute to split, we traverse every value $v_*$ of $V_*$, and the sub-model can be obtained with the sub-dataset corresponding to $v_*$ recursively. Then, we attach the sub-model to the full model. Algorithm  \ref{FPQMCA} is the pseudocode.

\begin{algorithm}[h!]
\caption{Fast preceding questionnaire model creating algorithm (FPQMCA)
\newline \textbf{Input:} $D$, $V$, Index list: $L$, $\beta $
\newline \textbf{Output:} Fast preceding questionnaire model: $M$
}
\label{FPQMCA}
\begin{algorithmic}[1]
\State Create a node $M$
\label{FPQMCA_L1}
\State Let $N(L(0))$ be the number of 0s in $L$
\If{$N(L(0))==1$}
\State Let $V(L(0))$ be the investigation attribute whose corresponding index in $L$ is 1
\State \Return $M$ as a leaf node labeled with $V(L(0))$
\label{FPQMCA_L5}
\Else
\State $V_*=BASCA(D,V)$
\label{FPQMCA_L7}
\State Label node $M$ with $V_*$
\label{FPQMCA_L8}
\State Let $L(V^{-1}(V_*))$ be the corresponding index in $L$ of $V_*$
\State $L(V^{-1}(V_*))=1$
\label{FPQMCA_L10}
\For{each value $v_*$ of $V_*$}
\label{FPQMCA_L11}
\State Let $D_{v_*}$ be the set of data tuples in $D$ satisfying $v_*$ on $V_*$
\State $M_{v_*}=FPQMCA(D_{v_*},V,L)$
\State Attach the node $M_{v_*}$ to node $M$
\EndFor
\label{FPQMCA_L14}
\EndIf
\State \Return $M$
\end{algorithmic}
\end{algorithm}

$L=List(0)$, and $L$ is initialized with a zero vector. Then, the FPQMCA can be called to obtain the fast preceding questionnaire model $M$. $M=FPQMCA(D,V,L)$.

\subsection{Model used for real investigation algorithm (MURIA)}

With the created FPQM, the new person in the testing dataset can be investigated quickly. At the beginning of the real investigation, there is no information about the respondent; therefore, all we can do is ask about the attribute directly. After sufficient information has been accumulated, some investigation attributes can be inferred; otherwise, we continue asking about attributes.

Let $Ind$ be the index indicating whether the current investigation attribute is the top attribute. $Ind=0$ indicates that it is the top attribute, and $Ind=1$ indicates that it is not. If the current investigation attribute is the top attribute, there is no information about the respondent; therefore, the attribute cannot be predicted.

$R'$, $I$, $O$, and $P$ are global variables that have be defined in Definitions \ref{R'}, \ref{I}, \ref{O}, and \ref{P}, respectively. The pseudocode is shown in Algorithm  \ref{MURIA}.

\begin{algorithm}[h!]
\caption{Model used for real investigation algorithm (MURIA)
\newline \textbf{Input:} $R_i$, $V$, $M$, threshold of confidence level: $\sigma$, Index: $Ind$
\newline \textbf{Output:} Last node: $LN$
}
\label{MURIA}
\begin{algorithmic}[1]
\State Let $M(1)$ be the top investigation attribute index of $M$. Let $V(M(1))$ be the top investigation attribute of $M$. Let $R_i(M(1))$ be the real value of $R_i$ on $V(M(1))$. Let ${R_i}'(M(1))$ be the final value of ${R_i}'$ on V(M(1)). Let $I_i(M(1))$ be the indication value of $I_i$ on $V(M(1))$. Let $P_i(M(1))$ be the confidence level of $P_i$ on $V(M(1))$. Let $O_i(j_1)$ be the current attribute of $O_i$.
\label{MURIA_L1}
\If{$Ind==0$}
\State ${R_i}'(M(1))=R_i(M(1))$
\State $I_i(M(1))=0$
\State $P_i(M(1))=1$
\State $Ind=1$
\EndIf
\State $O_i (j_1 )=M(1)$
\State Let $M_s$ be the sub-model of $M$ when $\tilde{U_i}$ takes ${R_i}'(M(1))$.
\State Let $M_s (1)$ be the top investigation attribute index of $M_s$. Let $V(M_s (1))$ be the top investigation attribute of $M_s$. Let $R_i (M_s (1))$ be the real value of $R_i$ on $V(M_s (1))$. Let ${R_i}'(M_s (1))$ be the final value of ${R_i}'$ on $V(M_s (1))$. Let $I_i (M_s (1))$ be the indication value of $I_i$ on $V(M_s (1))$. Let $P_i (M_s (1))$ be the confidence level of $P_i$ on $V(M_s (1))$. Let $O_i (j_2 )$ be the next attribute of $O_i$. Let $W(M_s (1))$ be possible values on $M_s (1)$. Let $C(v(M_s (1))\mid {R_i}' (M(1)) )$ be the confidence level of $M_s (1)$ taking $v(M_s (1))$ under the condition of $M(1)$ taking ${R_i}' (M(1))$.
\label{MURIA_L10}
\State $C_*=\underset{v(M_s (1))\in W(M_s (1))}{max}C(v(M_s (1))\mid {R_i}' (M(1)) )$
\label{MURIA_L11}
\If{$C_*>\sigma $}
\label{MURIA_L12}
\State ${R_i}' (M_s (1))=v(M_s (1))$
\State $I_i (M_s (1))=1$
\State $P_i (M_s (1))= C_*$
\State $O_i (j_2 )=M_s (1)$
\Else
\State ${R_i}' (M_s (1))=R_i (M_s (1))$
\State $I_i (M_s (1))=0$
\State $P_i (M_s (1))= 1$
\State $O_i (j_2 )=M_s (1)$
\EndIf
\If{$M_s\ !=a\ tree$}
\State \Return $M_s$
\label{MURIA_L24}
\Else
\label{MURIA_L25}
\State \Return $MURIA(R_i,V,M_s,\sigma ,Ind)$
\EndIf
\label{MURIA_L27}
\end{algorithmic}
\end{algorithm}

$Ind=0$ is the initial value. We traverse every testing dataset in $R$ as $R_i$, and we call the algorithm $LN=MURIA(R_i,V,M,\sigma ,Ind)$ to obtain $R'$, $I$, $O$, and $P$.

\subsection{Model evaluation algorithm (MEA)}

Now, the FPQM should be evaluated to determine its performance. Various evaluation metrics are calculated by the model evaluation algorithm (MEA). First, $K$ can be calculated with $R$ and $R'$. Then, $AAR$, $AAR$, and $AF_\beta $ can be obtained with $K$ and $I$. A larger $AF_\beta $ indicates a better FPQM. Algorithm  \ref{MEA} is the pseudocode.

\begin{algorithm}[h!]
\caption{Model evaluation algorithm (MEA)
\newline \textbf{Input:} $R$, $R'$, $I$, $\beta $
\newline \textbf{Output:} $AAR$, $ARR$, ${AF}_\beta$
}
\label{MEA}
\begin{algorithmic}[1]
\For{$i\leftarrow 1$ to $\tilde{m}$}
\label{MEA_L1}
\For{$j\leftarrow 1$ to $n$}
\If{${R_{ij}}'\neq R_{ij}$}
\State $K_{ij}=0$
\Else
\State $K_{ij}=1$
\EndIf
\EndFor
\EndFor
\label{MEA_L9}
\State Let Sum1, Sum2, Sum3, and Sum4 be temporary variables
\label{MEA_L10}
\State $Sum1=0$
\State $Sum2=0$
\State $Sum3=0$
\label{MEA_L12}
\For{$i\leftarrow 1$ to $\tilde{m}$}
\label{MEA_L13}
\State $Sum4=0$
\State $Sum5=0$
\For{$j\leftarrow 1$ to $n$}
\State $Sum4=Sum1+K_{ij}*I_{ij}$
\State $Sum5=Sum2+I_{ij}$
\EndFor
\State $AR_i=Sum4/Sum5$
\State $RR_i=Sum5/n$
\State $F_i= \frac{(\beta ^2+1)*AR_i*RR_i}{\beta ^2*AR_i+RR_i} $
\State $Sum1= Sum1+AR_i$
\State $Sum2= Sum2+RR_i$
\State $Sum3= Sum3+F_i$
\EndFor
\label{MEA_L24}
\State $AAR= Sum1/\tilde{m}$
\label{MEA_L25}
\State $ARR= Sum2/\tilde{m}$
\State $AF_\beta=Sum3/\tilde{m}$
\label{MEA_L27}
\State \Return $AAR$, $ARR$, $AF_\beta $
\end{algorithmic}
\end{algorithm}

When $\beta $ is given, $(AAR, ARR, AF_\beta )=MEA(R,R',I,\beta )$. $AAR$, $ARR$, $AF_\beta $ can be obtained by calling the MEA.

\subsection{An example}

Here is an example to illustrate the Definitions \ref{def1}-\ref{d5} and Algorithms \ref{BASCA}-\ref{MEA}.

\begin{table}[h]\centering
\caption{Example Training Dataset}\label{Training Dataset}
\begin{tabular}{cccccc}
\toprule
Investigation attributes& Education & Income & Social Skills & Work Ability & Communication\\
\midrule
$U_1$ & 0&1&0&1&1 \\
$U_2$& 1 &2 &0 &0 &1\\
$U_3$& 1&0 &1& 0& 1 \\
$U_4$& 1& 0 &1 &1& 0\\
\bottomrule
\end{tabular}
\end{table}

\begin{table}[h]\centering
\caption{Example Testing Dataset}\label{Testing Dataset}
\begin{tabular}{cccccc}
\toprule
Investigation attributes& Education & Income & Social Skills & Work Ability & Communication\\
\midrule
$\tilde{U_1}$ & 1 & 1 & 0 & 1 & 0 \\
$\tilde{U_2}$& 0 & 1 & 1 & 0 &  1\\
\bottomrule
\end{tabular}
\end{table}

$U=\{U_1,U_2,U_3,U_4,U_5\}$

$V=\{V_1,V_2,V_3,V_4,V_5\}=\{Education,Income,Social\ Skills,Work\ Ability,Communication\}$

$W=\{W_1,W_2,W_3,W_4,W_5\}=\{\{0,1\},\{0,1,2\},\{0,1\},\{0,1\},\{0,1\}\}$

$N=\{N_1,N_2,N_3,N_4,N_5\}=\{2,3,2,2,2\}$

$
D = \begin{bmatrix}
0&1&0&1&1 \\
1 &2 &0 &0 &1\\
1&0 &1& 0& 1 \\
1& 0 &1 &1& 0
     \end{bmatrix}
$

When $t=1$, no values of the investigation attributes are known yet.
By Eq.~\eqref{eq1}, 

\begin{equation}
\label{exameq1}
P(V_2^1=0|V_1^1=0)=\frac{N(V_2^1=0,V_1^1=0)}{N(V_1^1=0)}=\frac{0}{1}=0
\end{equation}

\begin{equation}
\label{exameq2}
P(V_2^1=1|V_1^1=0)=\frac{N(V_2^1=1,V_1^1=0)}{N(V_1^1=0)}=\frac{1}{1}=1
\end{equation}

\begin{equation}
\label{exameq3}
P(V_2^1=2|V_1^1=0)=\frac{N(V_2^1=2,V_1^1=0)}{N(V_1^1=0)}=\frac{0}{1}=0
\end{equation}

By Eq.~\eqref{eq2}, 

\begin{align}
\label{exameq4}
P(V_2^1|V_1^1=0)  &=P(V_2^1=0|V_1^1=0)^2+P(V_2^1=1|V_1^1=0)^2+P(V_2^1=2|V_1^1=0)^2\nonumber \\
 &=0^2+1^2+0^2=1
\end{align}

Similar to Eqs~\eqref{exameq1}-\eqref{exameq4},

\begin{align}
\label{exameq5}
P(V_2^1|V_1^1=1) &=P(V_2^1=0|V_1^1=1)^2+P(V_2^1=1|V_1^1=1)^2+P(V_2^1=1|V_1^1=1)^2 \nonumber \\
 &=\frac{2}{3}^2+0^2+\frac{1}{3}^2=\frac{5}{9}
\end{align}

By Eq.~\eqref{eq3},

\begin{align}
\label{exameq6}
P(V_2^1|V_1^1) &=P(V_1^1=0)*P(V_2^1|V_1^1=0)+P(V_1^1=1)*P(V_2^1|V_1^1=1)\nonumber \\
 &=\frac{1}{4}*1+\frac{3}{4}*\frac{5}{9}=\frac{2}{3}
\end{align}

Similar to Eqs~\eqref{exameq1}-\eqref{exameq6}, 

\begin{equation}
\label{exameq7}
\begin{split}
P(V_3^1|V_1^1)=\frac{2}{3}\quad 
P(V_4^1|V_1^1)=\frac{2}{3}\quad 
P(V_5^1|V_1^1)=\frac{2}{3}
\end{split}
\end{equation}

By Eq.~\eqref{eq4}, 

\begin{align}
\label{exameq8}
P(\Delta |V_1^1) &=P(V_2^1|V_1^1)+P(V_3^1|V_1^1)+P(V_4^1|V_1^1)+P(V_5^1|V_1^1)\nonumber \\
 &=\frac{2}{3}+\frac{2}{3}+\frac{2}{3}+\frac{2}{3}=\frac{8}{3}
\end{align}

Similar to Eqs~\eqref{exameq1}-\eqref{exameq8}, 

\begin{equation}
\label{exameq9}
\begin{split}
P(\Delta |V_2^1)=\frac{7}{2} &\quad 
P(\Delta |V_3^1)=\frac{11}{4}\\
P(\Delta |V_4^1)=\frac{5}{2}&\quad 
P(\Delta |V_5^1)=\frac{5}{2}
\end{split}
\end{equation}

By Eq.~\eqref{eq5}, 

\begin{align}
\label{exameq10}
V_*^1=\underset{V_{j_{k_1}}^1\in V}{arg\ max }\ INF(\Delta |V_{j_{k_1}}^1)=V_2^1
\end{align}

Eqs~\eqref{exameq1}-\eqref{exameq10} also shows how Algorithm \ref{BASCA} performs.

As for Algorithm \ref{FPQMCA}, a node M is created at the beginning. $L$ is initialized as a zero vector: $L=List(0)$. $N(L(0))=n$ and not 1, and $V_*^1=BASCA(D,V)$ with Algorithm \ref{BASCA}. We label node M as $V_*^1$, $L(V^{-1}(V_*^1))=1$.  $V_*^1=V_2^1$ (\texttt{Income}) and $W_2=\{0,1,2\}$; therefore, there are three branches from the attribute \texttt{Income}. On the branch $Income=0$, $D_{v_*} = \begin{bmatrix}
       1 & 1 & 0 & 1 \\
       1 & 1 & 1 & 0 
     \end{bmatrix}$, $M_{v_*}=FPQMCA(D_{v_*},V,L)$. Attach the node $M_{v_*}$ to node $M$. When $N(L(0))==1$, Algorithm \ref{FPQMCA} terminates, and  the fast preceding questionnaire model is created. Notice that only the first recursion step is shown here to ensure that the example is not too long.

\begin{figure}[ht]
    \centering
    \includegraphics[width=0.8\textwidth]{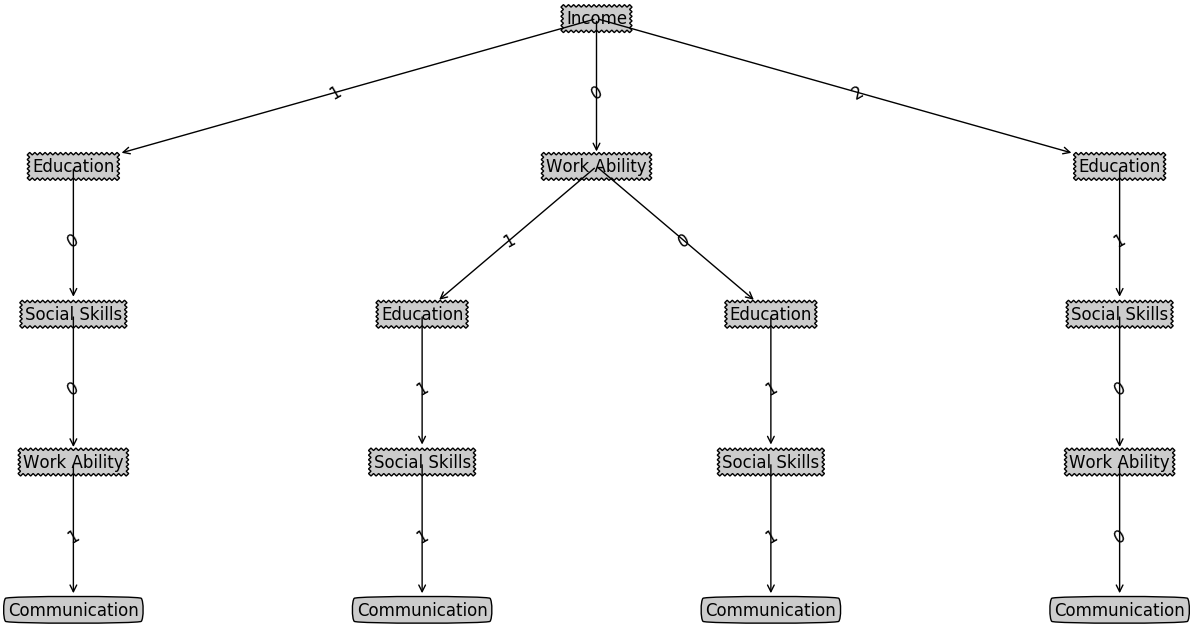}
    \caption{The obtained FPQM for the example training dataset.}
    \label{fig:exampleTree}
\end{figure}

The created model is shown in Figure \ref{fig:exampleTree}.

$\tilde{U}=\{\tilde{U_1},\tilde{U_2}\}$,
$
R = \begin{bmatrix}
       1 & 1 & 0 & 1 & 0 \\
       1 & 0 & 1 & 1 &  0
     \end{bmatrix}
$.

After the model is created, Algorithm  \ref{MURIA} can be used for a real investigation on the testing dataset. $V_*^1=V_2^1$; therefore, $M(1)=2$ and $V(M(1))=V_2^1$. $
R_1=\begin{bmatrix}
       1 & 1 & 0 & 1 & 0
     \end{bmatrix}
$. $Ind=0$, then ${R_1}'(M(1))={R_1}'(2)=R_1(2)=1$, $I_1(2)=0$, $P_1(2)=1$, $Ind=1$. $O_1(1)=2$. $M_s (1)=1$, $V(M_s(1))=V_1^1$.
$C_*=\underset{v(M_s (1))\in W(M_s (1))}{max}C(v(M_s (1))\mid {R_1}'(2))=1$ and $v(M_s (1))=0$. $C_*>\sigma =0.8$, then ${R_1}' (1)=0$, $I_1(1)=1$, $P_1(1)= 1$, $O_1 (2)=1$. $M_s=$ a tree; therefore, return $MURIA(R_1,V,M_s,\sigma ,1)$. When Algorithm  \ref{MURIA} terminates, we obtain $
R_1'=\begin{bmatrix}
      0& 1& 0& 1& 1
     \end{bmatrix}
$, $P_1$, $I_1$, and $O_1$.

After Algorithm  \ref{MURIA} finishes investigating the testing dataset $R$, we can obtain

$
R'=\begin{bmatrix}
      0& 1& 0& 1& 1\\
      1& 0& 1& 1& 0
     \end{bmatrix}$, $
P=\begin{bmatrix}
      1& 1& 1& 1& 1\\
      1& 1& 1& 0.5& 1
     \end{bmatrix}$, $
I=\begin{bmatrix}
      1& 0& 1& 1& 1\\
      1& 0& 1& 0& 1
     \end{bmatrix}$, $
O=\begin{bmatrix}
      2& 1& 3& 4& 5\\
      1& 3& 0& 2& 4
     \end{bmatrix}$.
     
    $M$ is the fast preceding questionnaire model, $C$ is the collection of all confidence levels that appear when creating $M$. $S=\{U,V,W,D,R,R',K,P,I,O,M,C\}$, where $\bar{S}$ is the space of all possible questionnaire models.
    
$AAR$, $ARR$, and $AF_\beta $ can be calculated by Algorithm  \ref{MEA}, which can be applied for model evaluation. With Lines \ref{MEA_L1}-\ref{MEA_L9}, $
K=\begin{bmatrix}
      1& 1& 1& 1& 0\\
      1& 1& 1& 1& 1
     \end{bmatrix}$. With Lines \ref{MEA_L10}-\ref{MEA_L27}, $AAR=0.7$, $ARR=0.75$, and $AF_{0.5} =0.7114$ with $\beta = 0.5$.

    The problem is defined as the following.
    
    \begin{flalign}
        max\ AF_\beta \ \ subject\ to\ S\in \bar{S} \nonumber
    \end{flalign}

\subsection{The comparison with Decision Tree}

The fast preceding questionnaire model (FPQM) is similar to Decision Tree to some extent, but they are different models. There are many types of Decision Tree algorithms. The notable models include ID3, C4.5, C5.0, CART, CHAID, MARS,  and Conditional Inference Trees. ID3, C4.5 and CART are chosen as representative Decision Tree algorithms. The FPQM is compared with these three Decision Tree algorithms in all aspects\cite{cinaroglucomparison,han2011data,singh2014comparative}. 

\subsubsection{The comparison on measures}
The algorithms for constructing decision trees choose an attribute at each step that best splits the dataset. Different Decision Tree algorithms use different metrics for determining the "best". ID3 uses information gain, C4.5 uses a gain ratio, and CART uses the Gini index, whereas the FPQM uses Influence, which is defined in Definitions \ref{d1}-\ref{d5}.

\subsubsection{The comparison on solved problems}
ID3, C4.5, and CART can solve both classification and regression problems. On the other hand, the FPQM is designed to optimize questionnaires by reordering the attributes and reducing the number of attributes by predicting the low-ranking attributes. The solved problems between the FPQM and Decision Tree are entirely different.

\subsubsection{The comparison on model forms}
The model form of the FPQM looks very similar to that of Decision Tree algorithms; both are in tree forms, as shown in Figure \ref{fig:wholeresult}.

\subsubsection{The comparison on attribute types}
The attribute types that the FPQM and Decision Tree can handle are different. ID3 can only handle nominal types, C4.5 and CART can handle both nominal and numeric types, and the FPQM can only handle nominal attribute types.

\subsubsection{The comparison on splits}
The split methods of the FPQM and Decision Tree are different. ID3 and C4.5 split the data in multiple ways, CART splits the data as a 2-way split, and the FPQM splits the data in multiple ways. CART can only create binary trees, whereas ID3, C4.5 and the FPQM can create general trees.

The comparisons between the FPQM with Decision Tree in all aspects are listed in Table \ref{ComparisonDT}.

\begin{table}[h]\centering
\caption{The comparison with Decision Tree}\label{ComparisonDT}
\begin{tabular}{ccccc}
\toprule
 & \multicolumn{3}{c}{Decision Tree} \\ 
\cmidrule(r){2-4}
Aspect & ID3& C4.5 & CART &FPQM\\
\midrule
Measure&Information gain& Gain ratio& Gini index&Influence\\
Solved problem&Classification&Classification&Classification/Regression&Reorder \& Reduce\\
Model form&Tree& Tree&Tree&Tree\\
Attribute type&Nominal&Nominal/Numeric&Nominal/Numeric&Nominal\\
Split&Multi-way&Multi-way&2-way&Multi-way\\
\bottomrule
\end{tabular}
\end{table}

\subsection{The time complexity of the FPQM}
There are many loops and recursive steps in the four algorithms constituting the FPQM; the time complexity of the FPQM is analyzed below. The derivation process of the time complexity is described in detail in Appendices.

\subsubsection{The time complexity of the BASCA}

The time complexity of the BASCA is 

\begin{equation}
\label{tBASCA}
T_{BASCA}(n,\bar{N})=O(n^2{\bar{N}}^2)
\end{equation}
where $\bar{N}$ is the average of $N_j$ in $N$, $\bar{N}=\frac{1}{n}\sum_{j=1}^{n}N_j$, and $n$ is the number of $V$.

\subsubsection{The time complexity of the FPQMCA}

The time complexity of the FPQMCA is 

\begin{equation}
\label{tFPQMCA}
\begin{split}
T_{FPQMCA}(n,\bar{N})=O(\bar{N}^{n+1}+n^2\bar{N})
\end{split}
\end{equation}

\subsubsection{The time complexity of the MURIA}
The time complexity of the MURIA is 

\begin{equation}
\label{r2MURIA}
\begin{split}
T_{MURIA}(n,\bar{N})=O(n\bar{N})
\end{split}
\end{equation}

\subsubsection{The time complexity of the MEA}
The time complexity of the MEA is 

\begin{align}
\label{tBASCA}
T_{MEA}(n,\tilde{m})=O(\tilde{m}n)
\end{align}
where $\tilde{m}$ is the number of $\tilde{U}$.

\subsubsection{The time complexity of the FPQM}
The time complexity of the FPQM is 

\begin{align}
\label{FPQM}
T_{FPQM}(n,\bar{N},\tilde{m})&=T_{BASCA}(n,\bar{N})+T_{FPQMCA}(n,\bar{N})+T_{MURIA}(n,\bar{N})+T_{MEA}(n,\tilde{m})\nonumber\\
&=O(n^2{\bar{N}}^2)+O(\bar{N}^{n+1}+n^2\bar{N})+O(n\bar{N})+O(\tilde{m}n)\nonumber\\
&=O(n^2{\bar{N}}^2)+O(\bar{N}^{n+1})+O(\tilde{m}n)
\end{align}

\section{Experiments}
\label{experiments}
\subsection{Experimental data}
The experiments are based on actual data from the Lime Family Limited Company (\emph{Lime Family}). \emph{Lime Family} focuses on the pension service field and is the nationwide leading
provider of high-quality home care services. \emph{Lime Family} originally required 45 investigation attributes to assess a newcomer. The investigation attributes are based on the Barthel Index \cite{mahoney1965functional} and an ability assessment for elderly adults \cite{wang2014ability}. A total of 45 investigation attributes seems excessive for the elderly individuals who are asked to give the value of the attributes one by one, and the order of the attributes is not reasonable. Therefore, every assessment requires approximately 15-20 minutes, which is also excessive, resulting in customer churn in practice.

\subsubsection{Noise}
There are three main classes of noise: spurious readings, measurement error, and background data. Some values of the \texttt{Height} and \texttt{Weight} attributes are 0, and some values of the \texttt{Age} attribute are 2015. These are the spurious reading type of noise data; it is easy for people to see that they are incorrect values. \texttt{Height}, \texttt{Weight} and \texttt{Age} are numeric attributes, and thus, these noise data are replaced with a sample mean.

\subsubsection{Missing values}
Certain values of the \texttt{Religion}, \texttt{Ground Walking}, and \texttt{Up Down Stairs} attributes are missing. \texttt{Religion} is a nominal attribute, and \texttt{Ground Walking} and \texttt{Up Down Stairs} are ordinal attributes; thus, these missing values are replaced with a sample mode.

After preprocessing, including discretization, handling missing values, addressing noise data, and textual data processing, all data are converted into nominal data. The investigation attributes can be observed in Table \ref{tab:Assessattributes}.

\begin{table}[h]\centering
\caption{Investigation attributes}\label{tab:Assessattributes}
\begin{tabular}{cccccc}
\toprule
No. & Attribute Name & No. & Attribute Name & No. & Attribute Name\\
\midrule
\rowcolor[gray]{.9}
0 & Sex & 15 & Transfer Bed Chair & 30 & Financial Affairs Capability\\
\hline
1 & Age & 16 & Ground Walking& 31 & Cognitive Function\\
\hline
\rowcolor[gray]{.9}
2 & Ethnic& 17 & Up Down Stairs & 32 & Attacks\\
\hline
3 & Education & 18 &Witted& 33 & Depressive Symptoms\\
\hline
\rowcolor[gray]{.9}
4 & Religion & 19 & Height& 34 & Consciousness Level\\
\hline
5 & Marital Status& 20 & Weight& 35 & Vision\\
\hline
\rowcolor[gray]{.9}
6 & Housing Condition& 21 & Glasses& 36 & Hearing\\
\hline
7 & Income& 22 & Hearing Aid& 37 & Communication\\
\hline
\rowcolor[gray]{.9}
8 & Eating & 23 & Go Shopping& 38 & Life Skills\\
\hline
9 & Bathe & 24 & Outings & 39 & Work Ability\\
\hline
\rowcolor[gray]{.9}
10 & Modify & 25 & Food Cooking& 40 & Time Spatial Orientation\\
\hline
11 & Clothing & 26 & Maintain Housework & 41 & People Orientation\\
\hline
\rowcolor[gray]{.9}
12 & Stool Control & 27 & Washing Clothes & 42 & Social Skills\\
\hline
13 & Pee Control & 28 & Use Phone Ability & 43 & Event\\
\hline
\rowcolor[gray]{.9}
14 & Toilet & 29 & Medication & 44 & Occurrences Times\\
\bottomrule
\end{tabular}
\end{table}

The entire experiment is performed using the 64-bit Python language  on a MacBook Pro, with a 2.2 GHz Intel Core i7 CPU and 16 GB of 1600 MHz DDR3 memory.

\subsection{Evaluation metrics}
To evaluate the performance of the FPQM, the accuracy rate, average accuracy rate, standard deviation of the accuracy rate, reduction rate, average reduction rate, and standard deviation of the reduction rate are defined as follows.

\textbf{Accuracy rate (AR)}:

\begin{equation}
\label{eq6}
AR_i=\tfrac{\sum_{j=1}^{n}K_{ij}*I_{ij}}{\sum_{j=1}^nI_{ij}}
\end{equation}

$AR_i$ is the accuracy rate of individual $\tilde{U_i}$.

\textbf{Average accuracy rate (AAR)}:

\begin{equation}
\label{eq7}
AAR=\tfrac{1}{\tilde{m}}\sum_{i=1}^{\tilde{m}}\tfrac{\sum_{j=1}^{n}K_{ij}*I_{ij}}{\sum_{j=1}^nI_{ij}}
\end{equation}

$AAR$ is the average accuracy rate of $\tilde{U}$ and describes how accurate the model can be.

\textbf{Standard deviation of the accuracy rate (SAR)}:

\begin{equation}
\label{eq8}
SAR=\sqrt{\tfrac{1}{\tilde{m}}\sum_{i=1}^{\tilde{m}}(AR_i-AAR)^2}
\end{equation}

$SAR$ is the standard deviation of the accuracy rate and describes the volatility of the accuracy rate.

\textbf{Reduction rate (RR)}:

\begin{equation}
\label{eq9}
RR_i=\tfrac{\sum_{j=1}^{n}I_{ij}}{n}
\end{equation}

$RR_i$ is the reduction rate of $\tilde{U_i}$.

\textbf{Average reduction rate (ARR)}:

\begin{equation}
\label{eq10}
ARR=\tfrac{1}{\tilde{m}}\sum_{i=1}^{\tilde{m}}\tfrac{\sum_{j=1}^nI_{ij}}{n}
\end{equation}

$ARR$ is the average reduction rate of $\tilde{U}$ and describes how the model can accelerate the questionnaire.

\textbf{Standard deviation of the reduction rate (SRR)}:

\begin{equation}
\label{eq11}
SRR=\sqrt{\tfrac{1}{\tilde{m}}\sum_{i=1}^{\tilde{m}}(RR_i-ARR)^2}
\end{equation}

$SRR$ is the standard deviation of the reduction rate and describes the volatility of the reduction rate.

\textbf{$F_\beta $-Measure (F)}:

\begin{equation}
\label{eq12}
F_{\beta _i}=\tfrac{(\beta ^2+1)*AR_i*RR_i}{\beta ^2*AR_i+RR_i}
\end{equation}

$F_{\beta _i}$ is the $F_\beta $-Measure of $\tilde{U_i}$.

\textbf{Average $F_\beta $-Measure (AF)}:

\begin{equation}
\label{eq13}
AF_\beta =\tfrac{1}{\tilde{m}}\sum_{i=1}^{\tilde{m}}\tfrac{(\beta ^2+1)*AR_i*RR_i}{\beta ^2*AR_i+RR_i}
\end{equation}

$AF_\beta $ is the average $F_\beta $-Measure of $\tilde{U}$ and describes a balance between $AAR$ and $ARR$.

When $I_{ij}=0\ \forall i,j$, $ARR=0$ and $AAR=1$; therefore, $AF_\beta =0$. When $I_{ij}=1\ \forall i,j$, $ARR=1$, and both $AAR$ and $AF_\beta $ will be low. A proper $I_{ij}$ should be chosen to maximize $AF_\beta $ with proper $AAR$ and $ARR$. $AF_\beta $ describes a balance between $AAR$ and $ARR$. $\beta = 0.5$ is chosen in this experiment because $ARR$ is as important as $AAR$, and the goal is to improve $ARR$ and $AAR$ simultaneously.

\textbf{Standard deviation of $F_\beta $-Measure (SF)}:

\begin{equation}
\label{eq14}
SF=\sqrt{\tfrac{1}{\tilde{m}}\sum_{i=1}^{\tilde{m}}(F_{\beta _i}-AF_\beta )^2}
\end{equation}

$SF$ is the standard deviation of the $F_\beta $-Measure and describes the volatility of the $F_\beta $-Measure, which is the volatility of the overall performance.

\subsection{The overall results}

\begin{figure}[ht]
    \centering
    \includegraphics[width=\textwidth]{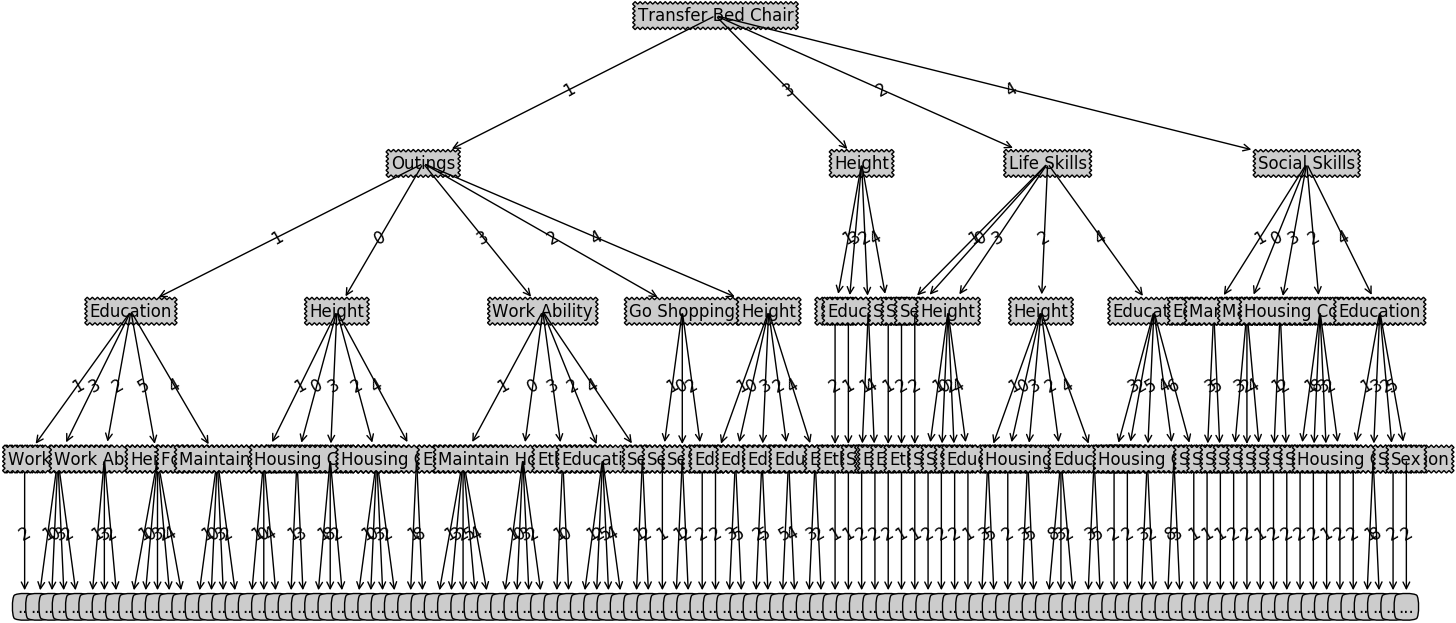}
    \caption{A part of the obtained FPQM.}
    \label{fig:wholeresult}
\end{figure}

The depth of the obtained FPQM is 45 because there are 45 investigation attributes in total. There is a corresponding rule with every node in the FPQM, and there are a total of 6072 rules. A part of the obtained FPQM is shown in Figure \ref{fig:wholeresult}.

A reasonable order for the investigation attributes can be obtained from the FPQM. \texttt{Transfer Bed Chair} is the first investigation attribute. When there is a new individual $\tilde{U_i}$ of $\tilde{U}$, if the \texttt{Transfer Bed Chair} value of $\tilde{U_i}$ equals 1, we enter the corresponding branch of the FPQM, and the second investigation attribute is \texttt{Outings}. This will continue recursively until the last investigation attribute and a reasonable order for $\tilde{U_i}$ are obtained during this process. The order will be as follows: \texttt{Transfer Bed Chair}, \texttt{Outings} etc. Every elderly man has his own exclusive reasonable questionnaire order, which is made especially for that individual \cite{kroh2016using}.

\begin{figure}[ht]
    \centering
    \includegraphics[width=0.8\textwidth]{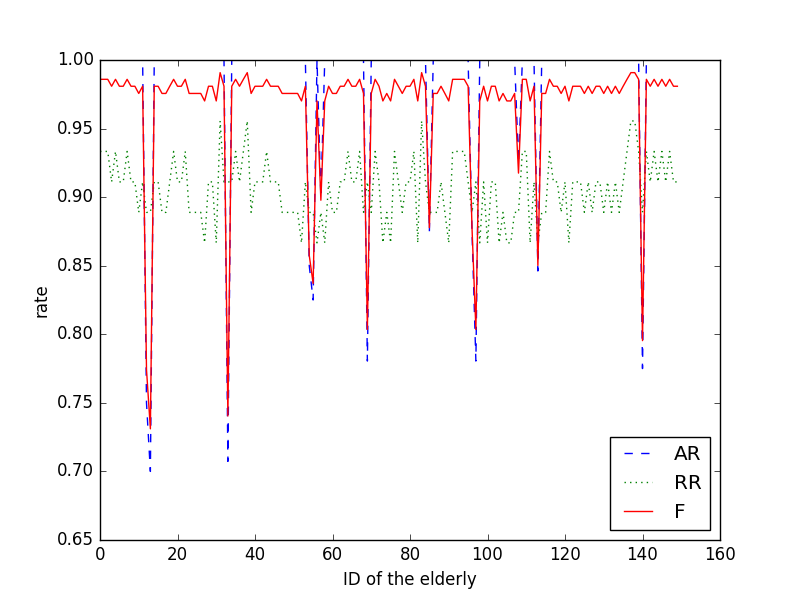}
    \caption{AR, RR and F of the FPQM.}
    \label{fig:FPQM}
\end{figure}

The FPQM is evaluated using the metrics $AR$, $RR$, and $F$, which are calculated with Eqs~\eqref{eq6},~\eqref{eq9}, and~\eqref{eq12}, respectively. The $AR$, $RR$, and $F$ of the FPQM are shown in Figure \ref{fig:FPQM}. The Y-axis indicates the values of $AR$, $RR$, and $F$, and the x-axis indicates the id of the elderly individual, thereby representing each investigated person. $AR$ is more volatile; accordingly, $F$ is also volatile. The FPQM is only a preceding questionnaire model, and the time required to obtain the pre-result is very short. To ensure 100\% accuracy, the obtained pre-result needs to be verified by the elderly individual, but verification is certainly much faster than asking directly.

\subsection{Comparison experiments}
To demonstrate that the FPQM is a good model for fast preceding questionnaires, Expert Knowledge, Rough Set\cite{liu2014three,gyorgy2003codecell}, and C4.5\cite{quinlan2014c4,han2011data}
are applied to solve the same problem. The results of the FPQM and those from these three methods are compared.

\subsubsection{Expert Knowledge}
\begin{figure}[ht]
    \centering
    \includegraphics[width=0.8\textwidth]{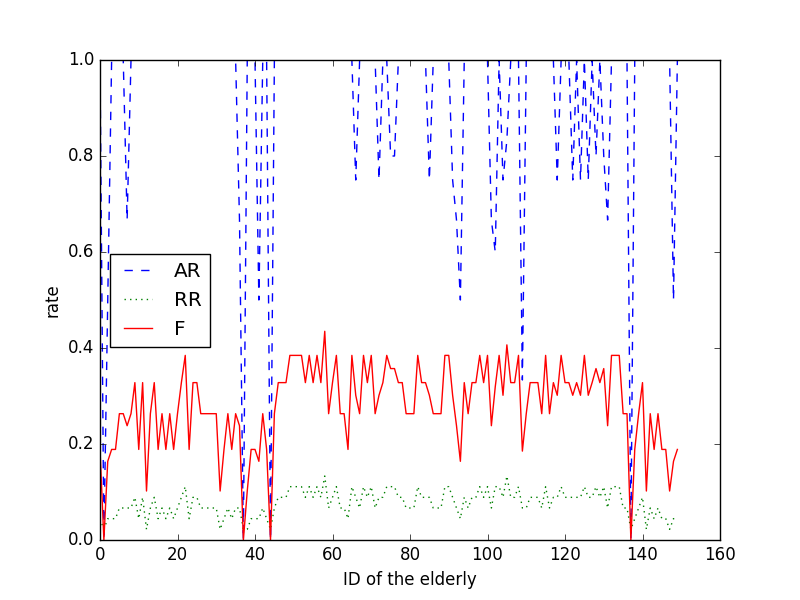}
    \caption{AR, RR and F of the Expert Knowledge method.}
    \label{fig:Expert_Knowledge}
\end{figure}

$AR$, $RR$ and $F$ are shown in Figure \ref{fig:Expert_Knowledge} when using the Expert Knowledge method. Expert Knowledge achieves a high $AR$, but the $RR$ is very low. This result conforms with the characteristics of experts, who ensure high $AR$ but do not greatly affect $RR$. $AR$ is more volatile than $RR$ between elderly individuals; accordingly, $F$ is also volatile.

\subsubsection{Rough Set}

The Rough Set method is also applied to solve the same problem in four steps. Step 1: First, take one attribute as the decision attribute and the other directions as the condition attributes. The correlation degree between each condition attribute and the decision attribute is calculated, and then, the attribute for which its correlation degree is less than a given threshold is deleted. Second, calculate the correlation degree between each condition with all the other conditions. Delete the condition attributes whose correlation degree with this particular condition attribute is larger than the degree with the decision attribute. Step 2: Generate the decision rules with the reliability and coverage degrees. Let the decision equivalent class (DEC) be the collection of elderly individuals who have the same assessment result. Calculate the reliability and coverage degree for each DEC and delete those classes whose degree is less than a given threshold. The classes that may generate rarely appearing rules and uncertain rules can be deleted in this manner. The decision rules can be created using each of the remaining decision equivalent classes. Step 3: Sort the investigation attributes with the coverage degree. Merge the rules according to the coverage degree and create the assessment sequence of attributes by sorting the coverage degrees in descending order. Step 4: Use the merged rules and assessment sequence to simulate the assessment process for the new elderly individual in $\tilde{U}$.

\begin{figure}[ht]
    \centering
    \includegraphics[width=0.8\textwidth]{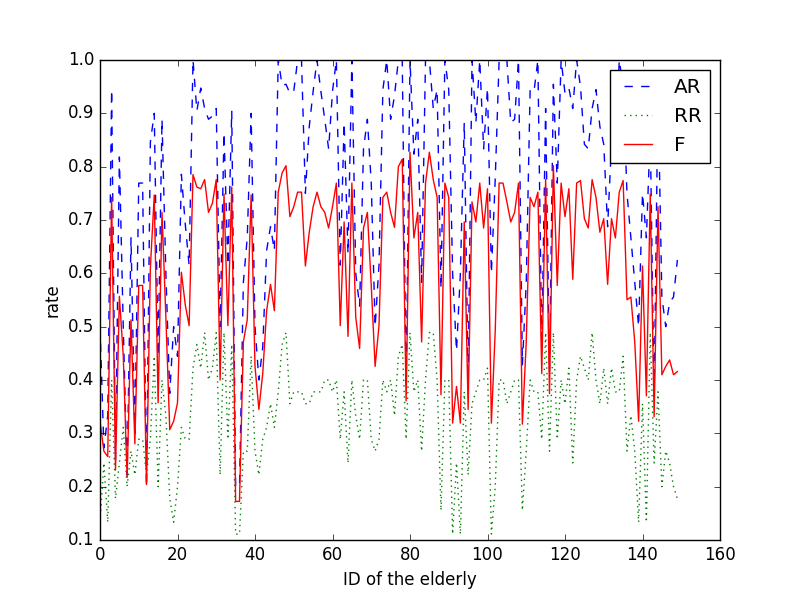}
    \caption{AR, RR and F of the Rough Set method.}
    \label{fig:Rough_Set}
\end{figure}

When using the Rough Set method, $AR$, $RR$ and $F$ are shown in Figure \ref{fig:Rough_Set}. Rough Set achieves a high $AR$, but $RR$ is slightly low. Both $AR$ and $RR$ are very volatile; therefore, $F$ is also quite volatile.

\subsubsection{C4.5}

Now, we solve the same problem by applying the C4.5 method. We choose each investigation attribute of the 45 attributes in total of V as the terminal node, and other attributes are internal nodes. Using the C4.5 method, a decision tree is obtained, and there are 45 decision trees with 45 attributes. Then, we evaluate the new individual $\tilde{U_i}$ of $\tilde{U}$ with the  45 obtained decision trees. Initially, there are no attributes that can be inferred, and we ask the elderly individual $\tilde{U_i}$ for the real value $R_{ij}$ directly. After accumulating sufficient real values $\{R_{ij}\}$,$j\in \{1,2,...,j_*\}$, some attributes can be inferred as the leaf nodes of some of the  45 obtained decision trees; then, we can infer their final values $R_{ij}'$. Otherwise, we continue to ask about $R_{ij}$ directly. Finally, all final values $R_{ij}'$ of the 45 attributes can be obtained. 

\begin{figure}[ht]
    \centering
    \includegraphics[width=0.8\textwidth]{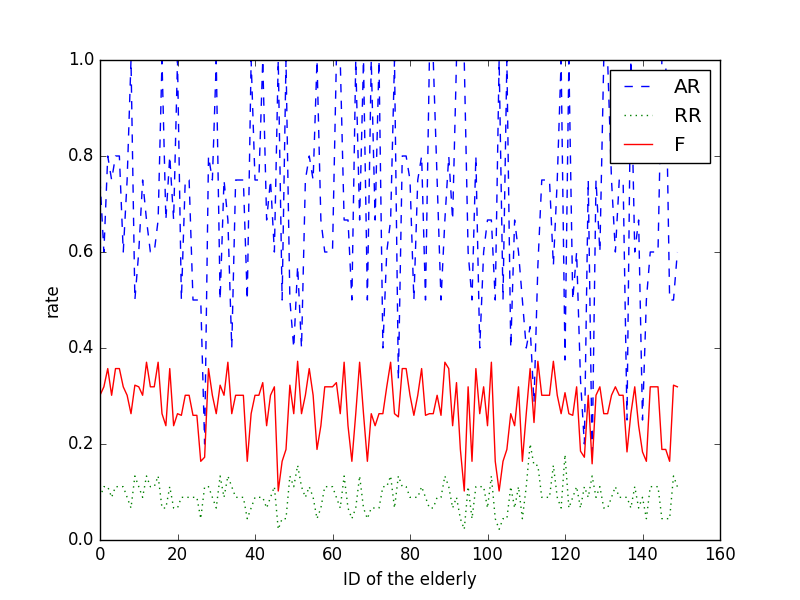}
    \caption{AR, RR and F of the C4.5 method.}
    \label{fig:C45}
\end{figure}

$AR$, $RR$ and $F$ using the C4.5 method are shown in Figure \ref{fig:C45}. C4.5 achieves very low $AR$, $RR$ and $F$. The volatility of $AR$ is slightly high, but C4.5 achieves a significantly lower volatility in $RR$ and $F$.

\subsubsection{Comparison results}

The FPQM, Expert Knowledge, Rough Set, and C4.5 methods are compared with respect to $AR$, $RR$ and $F$ individually to demonstrate which method achieves the best result.

\begin{figure}[ht]
    \centering
    \includegraphics[width=0.8\textwidth]{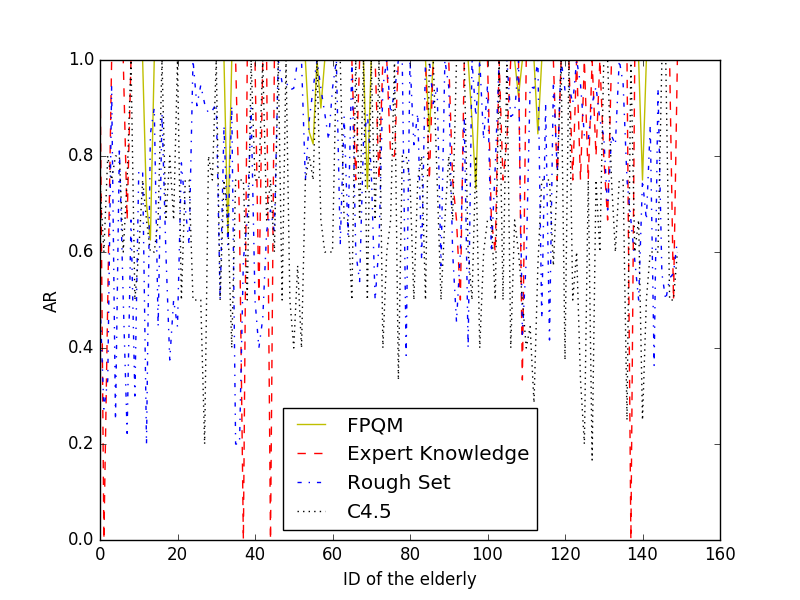}
    \caption{The AR comparison for the FPQM, Expert Knowledge, Rough Set, and C4.5 methods.}
    \label{fig:AR}
\end{figure}

Figure \ref{fig:AR} shows the $AR$ comparison for the FPQM, Expert Knowledge, Rough Set, and C4.5 methods. The FPQM achieves the best $AR$ performance, and the volatility is the lowest. Expert Knowledge achieves a slightly better $AR$ than the Rough Set and C4.5 methods, but these three methods have almost the same volatility.

\begin{figure}[ht]
    \centering
    \includegraphics[width=0.8\textwidth]{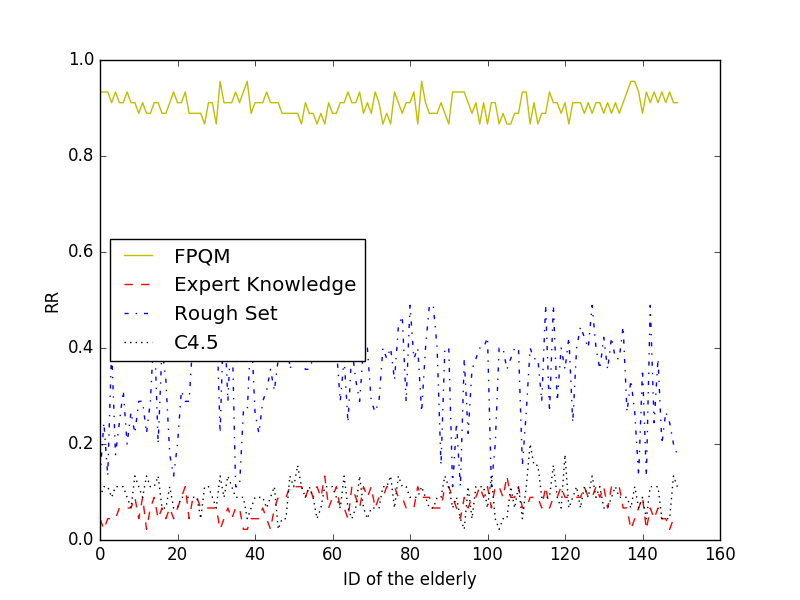}
    \caption{The RR comparison of the FPQM, Expert Knowledge, Rough Set, and C4.5 methods.}
    \label{fig:RR}
\end{figure}

The $RR$ comparison of the FPQM, Expert Knowledge, Rough Set, and C4.5 methods is shown in Figure \ref{fig:RR}. Apparently, the FPQM also achieves the best $RR$, which is far higher than those of the other three methods. The Rough Set method achieves the  second best value and has the greatest volatility. By comparison, $RR$ of the Expert Knowledge and C4.5 methods are the worst and approximately equal.

\begin{figure}[ht]
    \centering
    \includegraphics[width=0.8\textwidth]{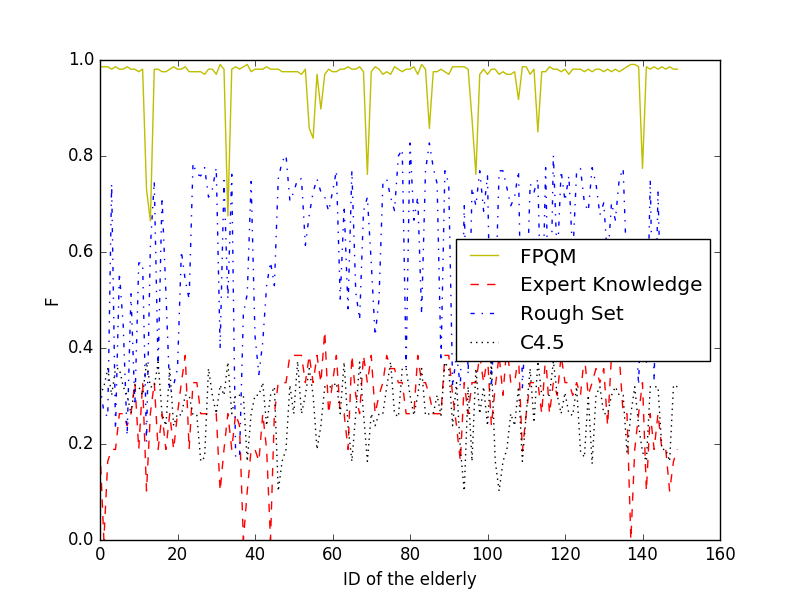}
    \caption{The F comparison of the FPQM, Expert Knowledge, Rough Set, and C4.5 methods.}
    \label{fig:F}
\end{figure}

$F$ is an integrated measurement of $AR$ and $RR$. Figure \ref{fig:F} shows the $F$ comparison for the FPQM, Expert Knowledge, Rough Set, and C4.5 methods. The FPQM achieves the best $AR$ and $RR$, as will $F$ naturally, and the volatility is the lowest. Because the other three methods have almost the same $AR$ volatility, the $F$ volatility is mainly determined by $RR$. The $F$ performance of the Rough Set method is the second best but is the most volatile. The  Expert Knowledge and C4.5 methods are the worst and approximately equal, but they do not exhibit a very high volatility.

Table \ref{tab:MeanSD} shows the mean and standard deviation of the four methods with respect to $AR$, $RR$, and $F$. The mean reflects the overall performance, and the standard deviation reflects the volatility in digital form. The mean of $AR$ is denoted as $AAR$, as mentioned previously; similar notation is used for $RR$ and $F$. These values can be calculated with Eqs~\eqref{eq7},~\eqref{eq10}, and~\eqref{eq13}. The standard deviation of $AR$ is denoted as $SAR$; similar notation is used for $RR$ and $F$. These values can be calculated with Eqs~\eqref{eq8},~\eqref{eq11}, and~\eqref{eq14}.

\begin{table}[!htbp]
\centering
\caption{Mean and standard deviation of the results under the four methods}\label{tab:MeanSD}
\begin{tabular}{cccc}
\hline
Metric& Method& Mean& Standard deviation\\
\hline
\multirow{4}*{AR}& FPQM& \textbf{0.9839}& \textbf{0.0560}\\
& Expert Knowledge& 0.9203& 0.1993\\
& Rough Set& 0.7589& 0.2223\\
& C4.5& 0.6850& 0.2072\\
\hline
\multirow{4}*{RR}& FPQM& \textbf{0.9056}& \textbf{0.0221}\\
& Expert Knowledge& 0.0769& 0.0262\\
& Rough Set& 0.3335& 0.1003\\
& C4.5& 0.0916& 0.0312\\
\hline
\multirow{4}*{F}& FPQM& \textbf{0.9666}& \textbf{0.0461}\\
& Expert Knowledge& 0.2800& 0.0870\\
& Rough Set& 0.5984& 0.1758\\
& C4.5& 0.2778& 0.0637\\
\hline
\end{tabular}
\end{table}

The FPQM has the largest mean and the smallest standard deviation with respect to $AR$, $RR$ and $F$. This means that the FPQM achieves the best performance and has the lowest volatility. The Expert Knowledge method has the second best mean for $AR$, 0.9203, and it has the smallest mean for $RR$, which conforms to the characteristics  of expert knowledge. The Rough Set method is the most volatile with respect to $AR$, $RR$ and $F$, and it achieves the second best mean for $F$ and $RR$. The performance of the C4.5 method is almost the worst.

\subsubsection{Result analysis}
Some of the improvements in the results are not highly significant when the FPQM is
compared with other methods, such as the Rough Set method. We can explain this as follows. The Rough Set method is also an excellent method for attribute reduction. Both the Rough Set and FPQM methods can capture the internal relationships among the attributes listed in Table \ref{tab:Assessattributes}, and predict multiple attributes simultaneously. Therefore, the FPQM does not achieve highly significant improvements compared with the Rough Set method.

\subsection{Factor analysis}
Three factors are analyzed: the number of elderly individuals, the number of investigation attributes, and the threshold $\sigma $. The evaluation metrics $AAR$, $ARR$, $AF$, $SAR$, $SRR$, and $SF$ are used to show the changes in the results. $AAR$, $ARR$, and $AF$ can be calculated with Eq.~\eqref{eq7},~\eqref{eq10}, and~\eqref{eq13}, respectively. Eq.~\eqref{eq8},~\eqref{eq11}, and~\eqref{eq14} can be used to calculate $SAR$, $SRR$, and $SF$, respectively.

\begin{figure}[ht]
    \centering
    \includegraphics[width=0.8\textwidth]{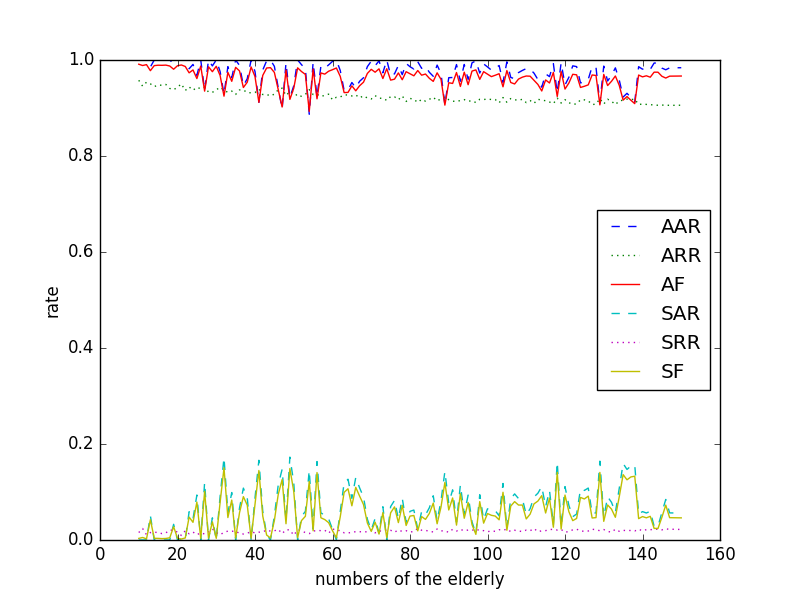}
    \caption{AAR, ARR, AF, SAR, SRR, and SF with increasing number of elderly individuals.}
    \label{fig:elder}
\end{figure}

As illustrated in Figure \ref{fig:elder}, $ARR$ decreases steadily and with minimal volatility when the number of the elderly individuals increases. $AAR$ and $AF$ are volatile and present a slight wavelike decrease overall. $SRR$ increases continuously with a small slope. $SF$ is highly similar to $SAR$ because $SRR$ is so small that $SF$ is mainly determined by $SAR$.

\begin{figure}[ht]
    \centering
    \includegraphics[width=0.8\textwidth]{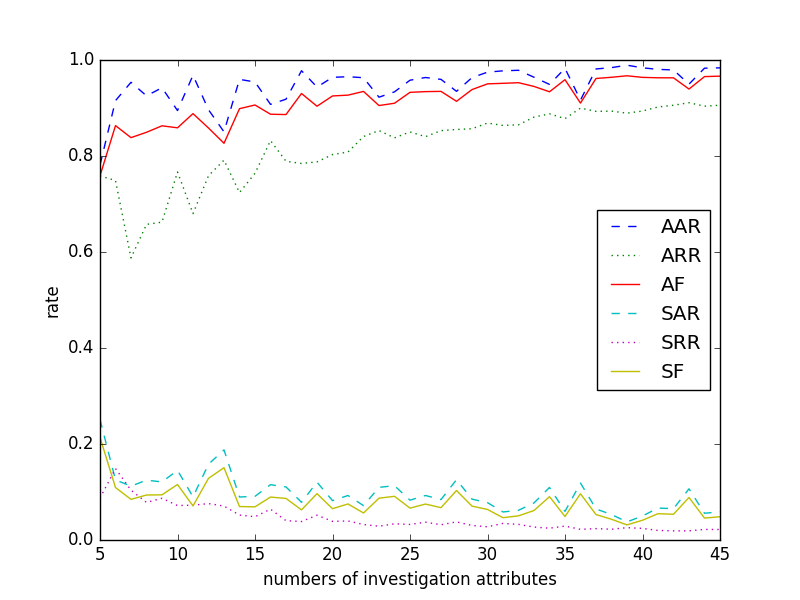}
    \caption{AAR, ARR, AF, SAR, SRR, and SF with increasing number of investigation attributes.}
    \label{fig:eval}
\end{figure}

With  increasing  number of investigation attributes,  $AAR$, $ARR$, and $AF$ present a fluctuating increasing trend, as shown in Figure \ref{fig:eval}. The increased level of $ARR$ is higher than that of $AAR$ and $AF$, and $ARR$ shows slightly larger volatility. $SAR$, $SRR$, and $SF$ decrease with some volatility, and $SRR$ falls fairly steadily in the second half of the curve. $SRR$ has minimal impact on $SF$, and $SF$ has the same general trend as $SAR$. Briefly, more investigation attributes produce better results.

\begin{figure}[ht]
    \centering
    \includegraphics[width=0.8\textwidth]{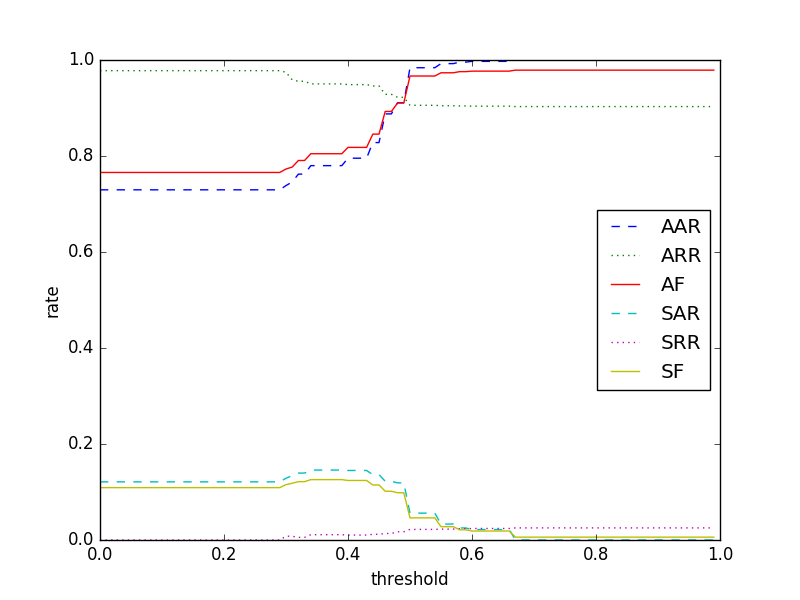}
    \caption{AAR, ARR, AF, SAR, SRR, and SF with increasing threshold $\sigma$.}
    \label{fig:threshold}
\end{figure}

Notice that some parts of the $AAR$, $ARR$, $AF$, $SAR$, $SRR$, and $SF$ curves are flat, which simply embodies the intermittent nature of the threshold, as illustrated in Figure \ref{fig:threshold}. Only the threshold $\sigma $ achieves a level whereby a significant change can occur. Changes mainly occur when $\sigma $ is in the range of 0.3-0.65. If $\sigma $ is too small, $I_{ij}=1\ \forall 1\leqslant i\leqslant \tilde{m}, 1\leqslant i\leqslant n$. On the other hand, $I_{ij}=0\ \forall 1\leqslant i\leqslant \tilde{m}, 1\leqslant i\leqslant n$ when $\sigma $ is too large. When the threshold $\sigma $ increases, $AAR$, $AF$, and $SRR$ present an increasing trend, whereas $ARR$, $SAR$ and $SF$ decrease. $SRR$ simply shows a slow growth. $ARR$ and $SRR$ have minimal impact on $AF$ and $SF$, respectively. Therefore, $AF$ and $SF$ show the same general trends as $AAR$ and $SAR$, respectively. Therefore, to achieve better results, it is suggested to choose a slightly larger threshold $\sigma $.

\section{Conclusion}
\label{conclusion}
Traditional questionnaires are used in a manner whereby respondents are asked one question after another. The main two problems are inefficiency and that the order of the questions is not reasonable. In this paper, the fast preceding questionnaire model (FPQM) is proposed to solve these problems in five steps, as shown in Figure \ref{fig:graph}. The influence calculation formula, best attribute to split choosing algorithm (BASCA), fast preceding questionnaire model creating algorithm (FPQMCA), model used for real investigation algorithm (MURIA),  and model evaluation algorithm (MEA) are all presented. The experimental section presents the experimental data; the evaluation metrics; the overall results of the FPQM; the comparison experiments with the Expert Knowledge, Rough Set, and C4.5 methods; and factor analysis, which includes the number of elderly individuals, the number of investigation attributes, and the threshold. When the FPQM is applied by the \emph{Lime Family} company, after asking about certain attributes, most of the remaining attributes could be inferred automatically with a high accuracy and reduction rate and low volatility. To ensure 100\% correctness, the model can be considered as a preceding questionnaire. Then, the elderly individual can perform verification, which is much faster than asking questions directly.

Further work will focus on three points. After the elderly individual is well assessed, a thorough study should be performed on determining an appropriate health care plan to be recommended automatically according to the assessment result. Then, the effect of the health care plan should be assessed as well. Moreover, as the amount data on elderly individuals continues increasing, certain distributed platforms, such as Hadoop and Spark, will be considered.

\acknowledgements
This work is supported by the National High Technology Research and Development Program (“863” Program) of China under Grant No. 2015AA016009 and the National Natural Science Foundation of China under Grant No. 61232005

The authors wish to thank Lei Yang from the \emph{Lime Family} Company, who provided us with the evaluation data for the elderly individuals.

Declaration: The data presented herein were only used for academic research. The customer ids have been anonymized, and the privacy of the elderly will not be invaded.

\bibliography{reference}{}
\bibliographystyle{ida}
\begin{appendices}
\label{appendices}
\section{The time complexity of the FPQM}

\subsection{The time complexity of the BASCA}
As shown in Algorithm  \ref{BASCA}, there are two loops on the attributes (Line \ref{BASCA_L1} and Line \ref{BASCA_L2}) and two loops on the values of the attributes (Line \ref{BASCA_L3} and Line \ref{BASCA_L4}) in the BASCA. As shown in Definition \ref{V}, $V=\{V_1,V_2,...,V_n\}$; therefore, there are n investigation attributes in $V$. $V_{j_{k_1}}^t$ has $N_{jk_1}$ possible values, where $N_{jk_1}$ is defined in Definition \ref{N}. Therefore, the time complexity of Lines \ref{BASCA_L1}-\ref{BASCA_L9} is 

\begin{equation}
\label{t1BASCA}
T_{BASCA}^1(n,\bar{N})=O(n^2{\bar{N}}^2)
\end{equation}where $\bar{N}$ is the average of $N_j$ in $N$, $\bar{N}=\frac{1}{n}\sum_{j=1}^{n}N_j$, and $n$ is the number of $V$.

Line \ref{BASCA_L10} calculates $V_*^t$ with Eq.~\eqref{eq5}, and  there are n investigation attributes in $V$; therefore, the time complexity of Line \ref{BASCA_L10} is 

\begin{equation}
\label{t2BASCA}
T_{BASCA}^2(n,\bar{N})=O(n)
\end{equation}

Thus, the time complexity of the BASCA is 

\begin{equation}
\label{tBASCA}
T_{BASCA}(n,\bar{N})=T_{BASCA}^1(n,\bar{N})+T_{BASCA}^2(n,\bar{N})=O(n^2{\bar{N}}^2)
\end{equation}

\subsection{The time complexity of the FPQMCA}
The time complexity of Lines \ref{FPQMCA_L1}-\ref{FPQMCA_L5} is 

\begin{equation}
\label{t1FPQMCA}
T_{FPQMCA}^1(n,\bar{N})=O(1)
\end{equation}

The time complexity of Line \ref{FPQMCA_L7} is 

\begin{equation}
\label{t2FPQMCA}
T_{FPQMCA}^2(n,\bar{N})=O(n^2{\bar{N}}^2)
\end{equation}

The time complexity of Lines \ref{FPQMCA_L8}-\ref{FPQMCA_L10} is 

\begin{equation}
\label{t3FPQMCA}
T_{FPQMCA}^3(n,\bar{N})=O(1)
\end{equation}

The time complexity of Lines \ref{FPQMCA_L11}-\ref{FPQMCA_L14} is 

\begin{equation}
\label{t4FPQMCA}
T_{FPQMCA}^4(n,\bar{N})=\bar{N}*T_{FPQMCA}(n-1)
\end{equation}

The time complexity of the FPQMCA is 

\begin{align}
\label{rFPQMCA}
T_{FPQMCA}(n,\bar{N})&=T_{FPQMCA}^1(n,\bar{N})+T_{FPQMCA}^2(n,\bar{N})+T_{FPQMCA}^3(n,\bar{N})+T_{FPQMCA}^4(n,\bar{N})\nonumber \\
&=O(1)+O(n^2{\bar{N}}^2)+O(1)+\bar{N}*T_{FPQMCA}(n-1,\bar{N})\nonumber \\
&=O(n^2{\bar{N}}^2)+\bar{N}*T_{FPQMCA}(n-1,\bar{N})
\end{align}

Eq.~\eqref{rFPQMCA} is the recursion formula of $T_{FPQMCA}(n,\bar{N})$.

\begin{align}
\label{r1FPQMCA}
T_{FPQMCA}(n,\bar{N})&=O(n^2{\bar{N}}^2)+\bar{N}*T_{FPQMCA}(n-1,\bar{N})\nonumber\\
&=O(n^2{\bar{N}}^2)+\bar{N}*\big\{O\big((n-1)^2\bar{N}^2\big)+\bar{N}*T_{FPQMCA}(n-2,\bar{N})\big\}\nonumber\\
&=O(n^2{\bar{N}}^2)+O\big((n-1)^2\bar{N}^3\big)+\bar{N}^2*T_{FPQMCA}(n-2,\bar{N})\nonumber\\
&=O(n^2{\bar{N}}^2)+O\big((n-1)^2\bar{N}^3)\big)+...+O(1^2\bar{N}^{n+1})+\bar{N}^{n-1}*T(1,\bar{N})\nonumber\\
&=\sum _{i=1}^nO(i^2{\bar{N}}^{n-i+2})+\bar{N}^{n-1}*T(1,\bar{N})\nonumber\\
&=O\big(\sum _{i=1}^n(i^2{\bar{N}}^{n-i+2})\big)+\bar{N}^{n-1}*T(1,\bar{N})
\end{align}

Let $S=\sum _{i=1}^n(i^2{\bar{N}}^{n-i+2})$.

\begin{equation}
\label{r2FPQMCA}
\begin{split}
(\bar{N}-1)S=\bar{N}^{n+2}+\sum _{i=1}^{n-1}\big((i+i+1){\bar{N}}^{n-i+2}\big)-n^2\bar{N}^2
\end{split}
\end{equation}

\begin{equation}
\label{r3FPQMCA}
\begin{split}
\bar{N}(\bar{N}-1)S-(\bar{N}-1)S=\bar{N}^{n+3}+2\sum _{i=4}^{n+2}\bar{N}^i-(n^2+2n-1)\bar{N}^3+n^2\bar{N}^2
\end{split}
\end{equation}

\begin{equation}
\label{r4FPQMCA}
\begin{split}
S=\frac{1}{(\bar{N}-1)^2}\big[\bar{N}^{n+3}+\frac{2\bar{N}^4(1-\bar{N}^{n-1})}{1-\bar{N}}-(n^2+2n-1)\bar{N}^3+n^2\bar{N}^2\big]
\end{split}
\end{equation}

Plugging Eq.~\eqref{r4FPQMCA} into Eq.~\eqref{r1FPQMCA},  $T(1,\bar{N})=O(1)$; thus,

\begin{align}
\label{r5FPQMCA}
&T_{FPQMCA}(n,\bar{N})\nonumber\\
&=O\Big(\frac{1}{(\bar{N}-1)^2}\big[\bar{N}^{n+3}+\frac{2\bar{N}^4(1-\bar{N}^{n-1})}{1-\bar{N}}-(n^2+2n-1)\bar{N}^3+n^2\bar{N}^2\big]\Big)+\bar{N}^{n-1}*T(1,\bar{N})\nonumber\\
&=O\Big(\frac{1}{(\bar{N}-1)^2}\big[\bar{N}^{n+3}+\frac{2\bar{N}^4(1-\bar{N}^{n-1})}{1-\bar{N}}-(n^2+2n-1)\bar{N}^3+n^2\bar{N}^2\big]\Big)+O(\bar{N}^{n-1})\nonumber\\
&=O\Big(\frac{1}{\bar{N}^2}\big[\bar{N}^{n+3}+\frac{\bar{N}^4\bar{N}^{n-1}}{\bar{N}}-n^2\bar{N}^3+n^2\bar{N}^2\big]\Big)+O(\bar{N}^{n-1})\nonumber\\
&=O(\bar{N}^{n+1}+n^2\bar{N})
\end{align}

The time complexity of the FPQMCA is 

\begin{equation}
\label{tFPQMCA}
\begin{split}
T_{FPQMCA}(n,\bar{N})=O(\bar{N}^{n+1}+n^2\bar{N})
\end{split}
\end{equation}

\subsection{The time complexity of the MURIA}
The time complexity of Lines \ref{MURIA_L1}-\ref{MURIA_L10} is 

\begin{equation}
\label{t1MURIA}
T_{MURIA}^1(n,\bar{N})=O(1)
\end{equation}

The time complexity of Line \ref{MURIA_L11} is 

\begin{equation}
\label{t2MURIA}
T_{MURIA}^2(n,\bar{N})=O(\bar{N})
\end{equation}

The time complexity of Lines \ref{MURIA_L12}-\ref{MURIA_L24} is 

\begin{equation}
\label{t3MURIA}
T_{MURIA}^3(n,\bar{N})=O(1)
\end{equation}

The time complexity of Lines \ref{MURIA_L25}-\ref{MURIA_L27} is 

\begin{equation}
\label{t4MURIA}
T_{MURIA}^4(n,\bar{N})=T_{MURIA}(n-1,\bar{N})
\end{equation}

The time complexity of the MURIA is 

\begin{align}
\label{rMURIA}
T_{MURIA}(n,\bar{N})&=T_{MURIA}^1(n,\bar{N})+T_{MURIA}^2(n,\bar{N})+T_{MURIA}^3(n,\bar{N})+T_{MURIA}^4(n,\bar{N})\nonumber\\
&=O(1)+O(\bar{N})+O(1)+T_{MURIA}(n-1,\bar{N})\nonumber\\
&=O(\bar{N})+T_{MURIA}(n-1,\bar{N})
\end{align}

Eq.~\eqref{rMURIA} is the recursion formula of $T_{MURIA}(n,\bar{N})$.

\begin{align}
\label{r1MURIA}
T_{MURIA}(n,\bar{N})&=O(\bar{N})+T_{MURIA}(n-1,\bar{N})\nonumber\\
&=2O(\bar{N})+T_{MURIA}(n-2,\bar{N})\nonumber\\
&=(n-1)O(\bar{N})+T_{MURIA}(1,\bar{N})
\end{align}

$T_{MURIA}(1,\bar{N})=O(1)$; therefore, the time complexity of the MURIA is 

\begin{equation}
\label{r2MURIA}
\begin{split}
T_{MURIA}(n,\bar{N})
=(n-1)O(\bar{N})+O(1)=O(n\bar{N})
\end{split}
\end{equation}

\subsection{The time complexity of the MEA}
The time complexity of Lines \ref{MEA_L1}-\ref{MEA_L9} is 

\begin{equation}
\label{t1MEA}
T_{MEA}^1(n,\tilde{m})=O(\tilde{m}n)
\end{equation}

The time complexity of Lines \ref{MEA_L13}-\ref{MEA_L24} is 

\begin{align}
\label{t2MEA}
T_{MEA}^2(n,\tilde{m})=O(\tilde{m}n)
\end{align}

The time complexity of Lines \ref{MEA_L10}-\ref{MEA_L12} and \ref{MEA_L25}-\ref{MEA_L27} is 

\begin{align}
\label{t3MEA}
T_{MEA}^3(n,\tilde{m})=O(1)
\end{align}

Thus, the time complexity of the MEA is 

\begin{align}
\label{tBASCA}
T_{MEA}(n,\tilde{m})=T_{MEA}^1(n,\tilde{m})+T_{MEA}^2(n,\tilde{m})+T_{MEA}^3(n,\tilde{m})=O(\tilde{m}n)
\end{align}

\subsection{The time complexity of the FPQM}
The time complexity of the FPQM is 

\begin{align}
\label{FPQM}
T_{FPQM}(n,\bar{N},\tilde{m})&=T_{BASCA}(n,\bar{N})+T_{FPQMCA}(n,\bar{N})+T_{MURIA}(n,\bar{N})+T_{MEA}(n,\tilde{m})\nonumber\\
&=O(n^2{\bar{N}}^2)+O(\bar{N}^{n+1}+n^2\bar{N})+O(n\bar{N})+O(\tilde{m}n)\nonumber\\
&=O(n^2{\bar{N}}^2)+O(\bar{N}^{n+1})+O(\tilde{m}n)
\end{align}

\end{appendices}
\end{document}